%% file: iclr2021_conference.tex
\documentclass{article} % For LaTeX2e
\usepackage{iclr2021_conference,times}

\input{math_commands.tex}

\usepackage{hyperref}
\usepackage{url}
\usepackage{graphicx}
\usepackage{comment}
\newtheorem{problem}{Problem}
\newtheorem{assumption}{Assumption}

\title{Safety Verification of Model Based \\ Reinforcement Learning Controllers} 

\author{Akshita Gupta \& Inseok Hwang \\
School of Aeronautics and Astronautics\\
Purdue University\\
\texttt{\{gupta417,ihwang\}@purdue.edu} 
}
\usepackage[ruled,vlined]{algorithm2e}
\SetKwInput{KwInput}{Input} 

\iclrfinalcopy % Uncomment for camera-ready version, but NOT for submission.
\begin{document}

\maketitle
\input{Abstract}
\begin{comment}
\begin{abstract}
The abstract paragraph should be indented 1/2~inch (3~picas) on both left and
right-hand margins. Use 10~point type, with a vertical spacing of 11~points.
The word \textsc{Abstract} must be centered, in small caps, and in point size 12. Two
line spaces precede the abstract. The abstract must be limited to one
paragraph.
\end{abstract}
\end{comment}

\input{introduction}
\input{Problem_formulation}
\input{Safety_verification}
%\input{methodology}
\input{experiments}

%\section{Experiments}
\section{Conclusion}
In this paper, we have presented a novel framework using forward and backward reachable tubes for safety verification and determination of the subset of initial states for which a given model-based RL controller always satisfies the state constraints.
%
%This complements various directions of research bring pursued to ensure safety of controllers, and can enrich the toolset for researchers and practitioners who aim to safely deploy model-based RL controllers to real physical systems.
%
%In the future, we aim to leverage the proposed framework to develop update procedures for searching safer controllers. 
 The main contribution of this work is the formulation of the reachability problem for a neural network modeled system dynamics and the use of level set method to compute an exact reachable tube by solving the Hamilton-Jacobi partial differential equation, for the reinforcement learning framework, thereby minimizing approximation errors that other existing reachability methods suffer.
 Additionally, the proposed framework can identify the set of safe initial sets for a given policy, thereby determining the initial conditions for which even a sub-optimal, unsafe policy satisfies the safety constraints.

\bibliography{iclr2021_conference}
\bibliographystyle{iclr2021_conference}

\clearpage
\appendix
\input{Appendix}

\end{document}

%% file: math_commands.tex
\usepackage{amsmath,amsfonts,bm}

\def\eqref#1{equation~\ref{#1}}
\def\1{\bm{1}}

\def\ra{{\textnormal{a}}}

\def\rx{{\textnormal{x}}}

\def\rva{{\mathbf{a}}}

\def\erva{{\textnormal{a}}}

\def\ervx{{\textnormal{x}}}

\def\rmA{{\mathbf{A}}}

\def\vmu{{\bm{\mu}}}
\def\vtheta{{\bm{\theta}}}
\def\va{{\bm{a}}}

\def\vd{{\bm{d}}}
\def\ve{{\bm{e}}}

\def\vo{{\bm{o}}}

\def\vs{{\bm{s}}}

\def\vx{{\bm{x}}}
\def\vy{{\bm{y}}}

\def\eva{{a}}

\def\mA{{\bm{A}}}

\def\mH{{\bm{H}}}
\def\mI{{\bm{I}}}
\def\mJ{{\bm{J}}}

\def\mW{{\bm{W}}}
\def\mX{{\bm{X}}}

\def\mSigma{{\bm{\Sigma}}}

\DeclareMathAlphabet{\mathsfit}{\encodingdefault}{\sfdefault}{m}{sl}
\SetMathAlphabet{\mathsfit}{bold}{\encodingdefault}{\sfdefault}{bx}{n}
\newcommand{\tens}[1]{\bm{\mathsfit{#1}}}
\def\tA{{\tens{A}}}

\def\tX{{\tens{X}}}

\def\gG{{\mathcal{G}}}

\def\sA{{\mathbb{A}}}
\def\sB{{\mathbb{B}}}
\def\sC{{\mathbb{C}}}
\def\sD{{\mathbb{D}}}
\def\sF{{\mathbb{F}}}

\def\sL{{\mathbb{L}}}

\def\sR{{\mathbb{R}}}
\def\sS{{\mathbb{S}}}
\def\sT{{\mathbb{T}}}

\def\emA{{A}}

\newcommand{\etens}[1]{\mathsfit{#1}}

\def\etA{{\etens{A}}}

\newcommand{\E}{\mathbb{E}}

\newcommand{\R}{\mathbb{R}}

\newcommand{\KL}{D_{\mathrm{KL}}}
\newcommand{\Var}{\mathrm{Var}}

\newcommand{\Cov}{\mathrm{Cov}}
\newcommand{\normltwo}{L^2}
\newcommand{\normlp}{L^p}

\newcommand{\parents}{Pa} % See usage in notation.tex. Chosen to match Daphne's book.

%% file: Abstract.tex
\begin{abstract}
    Model-based reinforcement learning (RL) has emerged as a promising tool for developing controllers for real world systems (e.g., robotics, autonomous driving, etc.).
    However, real systems often have constraints imposed on their state space which must be satisfied to ensure the safety of the system and its environment.
    %
    % The lack of a formal safety verification framework for RL algorithms hinders their deployment on safety critical systems.
    %
    Developing a verification tool for RL algorithms is challenging because the non-linear structure of neural networks impedes analytical verification of such models or controllers.
    %
    %
    %In addition to verifying the safety of a controller, a more useful result is the identification of initial conditions for which any given controller will satisfy the state constraints.
    %
    %
    To this end, we present a novel safety verification framework for model-based RL controllers using reachable set analysis.
    The proposed framework can efficiently handle models and controllers which are represented using neural networks.
    Additionally, if a controller fails to satisfy the safety constraints in general, the proposed framework can also be used to identify the subset of initial states from which the controller can be safely executed.
    %
    % Results are demonstrated on simulators motivated by real-world navigation problems.
\end{abstract}

%% file: introduction.tex
\section{Introduction}
%
\begin{comment}
Reinforcement learning (RL) paradigm has emerged as a powerful tool for developing optimal controllers for sequential decision making problems.
%
Particularly, RL algorithms don't assume \textit{a priori} knowledge of the system dynamics.
%
Instead, a data driven model of the system dynamics is developed either implicitly or explicitly.
%
This implies that any uncertainties in the system dynamics, like environment noise, friction etc., are also learnt by the modeled dynamics.
%
In this work, we narrow our focus on model-based RL algorithms with applications to safety critical physical systems.
\end{comment}

One of the primary reasons for the growing application of reinforcement learning (RL) algorithms in developing optimal controllers is that RL does not assume \textit{a priori} knowledge of the system dynamics.
Model-based RL explicitly learns a model of the system dynamics, from observed samples of state transitions.
This learnt model is used along with a planning algorithm to develop optimal controllers for different tasks.
Thus, any uncertainties in the system, including environment noise, friction, air-drag etc., can also be captured by the modeled dynamics.

However, the performance of the controller is directly related to how accurately the learnt model represents the true system dynamics.
Due to the discrepancy between the learnt model and the true model, the developed controller can behave unexpectedly when deployed on the real physical system, e.g., land robots, UAVs, etc. \citep{benbrahim1997biped, endo2008learning, morimoto2001acquisition}.
This unexpected behavior may result in the violation of constraints imposed on the system, thereby violating its safety requirements \citep{moldovan2012safe}.
Thus, it is necessary to have a framework which can ensure that the controller will satisfy the safety constraints \textit{before} it is deployed on a real system.
This raises the primary question of interest: \textit{Given a set of safety constraints imposed on the state space, how do we determine whether a given controller is safe or not?}

% While the RL framework doesn't organically allow for the incorporation of constraints, recent works have tried to address this problem.
%
In the literature, there have been several works that focus on the problem of ensuring safety.
Some of the works aim at \textit{finding} a safe controller, under the assumption of a known baseline safe policy \citep{hans2008safe,garcia2012safe,berkenkamp2017safe,thomas2015high,laroche2019safe}, or several known safe policies \citep{perkins2002lyapunov}. 
However, such safe policies may not be readily available in general.
Alternatively,  \citet{akametalu2014reachability} used reachability analysis to develop safe model-based controllers, under the assumption that the system dynamics can be modeled using Gaussian processes, i.e., an assumption which is violated by most modern RL methods that make use of neural networks (NN) instead.
%
% On the other hand, some works aim at \textit{evaluating} whether a given controller is safe or not using importance sampling \cite{doina}. 
%
% While importance sampling based methods do not model the system dynamics, their estimates of a controller's performance have variance exponential in the length of the trajectory \cite{guo}, 
% whereas works that use reachability analysis 
%
While there have been several works proposed to develop safe controllers, some of the assumptions made in these works may not be possible to realize in practice.
In recent years, this limitation has drawn attention towards developing \textit{verification} frameworks for RL controllers, which is the focus of this paper.
Since verifying safety of an NN based RL controller is also related to verifying the safety of the underlying NN model \citep{xiang2018reachable, tran2019parallelizable, xiang2018output, tran2019safety}, we provide an additional review for these methods in Appendix \ref{sec:ERW}. 

% . or controller.
% %
% NNs are one of the most commonly used function approximators for RL algorithms.
% %
% Therefore, often, verifying safety of an RL controller reduces to 
% %
% In this section, we briefly review the techniques used to evaluate the safety of NNs.
%

% Thus, there is an immediate need for the development of a safety verification framework which can handle NN based models for any given policy.
% \cite{akametalu2014reachability} focus on the problem of verifying whether

%
% However, some of these works make assumptions which may not be easy to realize in general.
% for all problem settings.
%
% \citet{hans2008safe} and \citet{garcia2012safe} assume that an initial safe policy is known before the controller is trained, while \citet{perkins2002lyapunov} switch between several safe policies during the learning process.
%
% \citet{berkenkamp2017safe} assumed that the safe starting policy can stabilize the system and proceeded to iteratively expand the safe set of states in the domain.
%
% \citet{akametalu2014reachability} used reachability analysis for the same purpose.
%
% It should be noted that in most of the works cited above, the system dynamics were modeled using Gaussian processes instead of neural networks (NNs) that are predominantly used in modern RL algorithms.
% , because the former is a more reliable statistical model.
%
% Thus, there is an immediate need for the development of a safety verification framework which can handle NN based models for any given policy.

%  by extending reachability analysis methods for NN based models and controllers. 
%

\paragraph{Contributions: }
In this work, we focus on the problem of determining whether a given controller is safe or not, with respect to satisfying constraints imposed on the state space.
%In this work, we focus on the latter problem of evaluating whether any given controller is safe or not.
%
%Specifically, we assume that there are state constraints imposed on the system, which must be avoided by the trained controller.
%
To do so, we propose a novel safety verification algorithm for model-based RL controllers using \textit{forward reachable tube} analysis that can handle NN based learnt dynamics and controllers, while also being robust against modeling error.
The problem of determining the reachable tube is framed as an optimal control problem using the Hamilton Jacobi (HJ) partial differential equation (PDE), whose solution is computed using the level set method.
%
% is the fact that they can represent sets with non-convex boundaries and that the results easily to extend to higher dimensions.
%
% Numerical schemes developed for level set methods are used to implicitly represent the reachable set by approximating the solution of the HJ PDE.
%
The advantage of using the level set method is the fact that it can represent sets with non-convex boundaries, thereby avoiding approximation errors that most existing methods suffer from.
Additionally, if a controller is deemed unsafe, we take a step further to identify if there are any starting conditions from which the given controller can be safely executed.
%
% Additionally, for a sub-optimal, unsafe controller we determine the subset of initial states, if any, from which the given controller can be safely executed without violating any constraints.
%
To achieve this, a \textit{backward reachable tube} is computed for the learnt model and, to the best of our knowledge, this is the first work which computes the backward reachable tube over an NN.
Finally, empirical results are presented on two domains inspired by real-world applications where safety verification is critical.

%% file: Problem_formulation.tex
\section{Problem Setting}
Let $\sS \subset \R^n$ denote the set of states and $\sA \subset \R^m$ denote the set of feasible actions for the RL agent.
Let $\sS_0 \subset \sS$ denote the set of bounded initial states and $\xi := \{ (\vs_t, \va_t) \}_{t=0}^T$ represent a trajectory generated over a finite time $T$, as a sequence of state and action tuples, where subscript $t$ denotes the instantaneous time.
Additionaly, let $\vs(\cdot)$ and $\va(\cdot)$ represent a sequence of states and actions, respectively.
% over some range whose endpoints are determined implicitly in a equation.
%
The state constraints imposed on the system are represented as unsafe regions using bounded sets $\sC_s = \cup_{i=1}^p \sC_s^{(i)}$, where $\sC_s^{(i)} \subset \sS, \; \forall i \in \{1, 2, \hdots p \}$. 
The true system dynamics is given by a non-linear function $f: \sS \times \sA \rightarrow \sR^n$ such that, $\dot{\vs} = f(\vs, \va)$, and is unknown to the agent.

A model-based RL algorithm is used to find an optimal controller $\pi: \sS \rightarrow \sA$, to reach a set of target states $\sT \subset \sS$ within some finite time $T$, while avoiding constraints $\sC_s$.
An NN model, $\hat{f}_\vtheta: \sS \times \sA \rightarrow \sR^n$ parameterized by weights $\vtheta$, is trained to learn the true, but unknown, system dynamics from the observed state transition data tuples $D = \{ ( \vs_t, \va_t, \Delta \vs_{t+1})^{(i)} \}_{i=1}^{N}$.
However, due to sampling bias, the learnt model $\hat{f}_\vtheta$ may not be accurate.
We assume that it is possible to estimate a bounded set $\sD \subset \sR^n$ such that, at any state $\vs \in \sS$, augmenting the learnt dynamics $\hat{f}_\vtheta$ with some $\vd \in \sD$ results in a closer approximation of the true system dynamics at that particular state.
%
%We assume that the modeling error is bounded, i.e., for a bounded set $\sD \subset \sR^n$,
%
%\begin{equation}
%    \forall \vs \in \sS, \; \exists \vd \in \sD \text{ such that, } f(\vs, \va) \approx \hat{f}_{\vtheta}(\vs, \va) + \vd.
%    \label{eqn:additive_model}
%\end{equation}
%
%The model in (\ref{eqn:additive_model}) is denoted as $\hat{f}^{(r)}_\vtheta: \sS \times \sA \times \sD \rightarrow \sR^n$.
%
Using this notation, we now define the problem of safety verification of a given controller $\pi(\vs)$.
\begin{problem} \label{prob1_safety_verification}
(Safety verification): Given a set of initial states $\sS_0$, determine if  $\,\, \forall \vs_0 \in \sS_0$, all the trajectories $\xi$ executed under $\pi(\vs)$ and following the system dynamics $f$, satisfy the constraints $\sC_s$ or not. %\todo{Rethink: true or estimated dynamics?}
\end{problem}
The solution to Problem \ref{prob1_safety_verification} will only provide a binary yes or no answer to whether $\pi(\vs)$ is safe or not with respect to $\sS_0$.
In the case where the policy is unsafe, a stronger result is the identification of safe initial states $\sS_{safe} \subset \sS_0$ from which $\pi(\vs)$ executes trajectories which always satisfy the constraints $\sC_s$.
This problem is stated below.
\begin{problem} \label{prob2_safe_states}
(Safe initial states): Given $\pi(\vs)$, find $\sS_{safe}$, such that, any trajectory $\xi$ executed under $\pi(\vs)$ and following the  system dynamics $f$, starting from any $\vs_0 \in \sS_{safe}$, satisfies the constraints $\sC_s$. %\todo{Rethink: true or estimated dynamics?}
\end{problem}
%
%The next section presents the proposed algorithm used to solve Problem \ref{prob1_safety_verification} and Problem \ref{prob2_safe_states} using reachability analysis.

%% file: Safety_verification.tex
\section{Safety Verification}
To address Problems \ref{prob1_safety_verification} and \ref{prob2_safe_states}, we use  \textit{reachability analysis}.
Reachability analysis is an exhaustive verification technique which tracks the evolution of the system states over a finite time, known as the \textit{reachable tube}, from a given set of initial states $\sS_0$ \citep{maler2008computing}.
If the evolution is tracked starting from $\sS_0$, then it is called the \textit{forward reachable tube} and is denoted as $\sF_R(T)$.
Analogously, if the evolution is tracked starting from $\sC_s$ to $\sS_0$, then it is called the \textit{backward reachable tube} and is denoted as $\sB_R(T)$.
%
% Depending on whether this evolution is tracked from or towards the set of initial states $\sS_0$, the reachable tube can be \textit{forward} that is denoted as $\sF_R(T)$, or \textit{backwards} that is denoted using $\sB_R(T)$, respectively.
%

In the following sections, we will formulate a reachable tube problem for NN-based models and controllers, and then propose a verification framework that (a) can determine whether or not a given policy $\pi$ is safe, and (b) can compute $\sS_{safe}$ if $\pi$ is unsafe.
% We first formulate the reachable sets for NN based model and controllers, and then develop a verification framework proposed in this work, the intersection of the reachable tube from $\sS_0$, with the constrained states $\sC_s$, is used to determine whether or not a given policy $\pi$ is safe.
%
% For an exhaustive verification technique, it is infeasible to sample trajectories from every point in the given initial state and further, to verify whether \textit{all} these trajectories satisfy the safety constraints.
% %
% Therefore, reachable sets are usually represented as polyhedrons and their evolution is tracked by pushing the boundaries of this polyhedron according to the system dynamics.
% %
% However, as discussed in Section 2, using convex polyhedrons for reachability analysis can lead to large approximation errors.
% %
% To avoid this, the level set methods are used to implicitly represent the boundaries of the reachable tube at every time instant.
% %
% Using this implicit representation, even non-convex boundaries of the reachable tube can be represented.
% %
% Since the non-linear, non-convex structure of NNs are not suitable for analytical verification techniques, the reachability analysis gives an efficient, simulation-based approach to analyze the safety of NNs.
% In the rest of the section we attempt to answer two questions.
To do so, there are two main questions that need to be answered.
First, since the true system dynamics $f$ is unknown, how can we determine a conservative bound on the modeling error, to augment the learnt model $\hat{f}_\vtheta$ and better model the true system dynamics when evaluating a controller $\pi$? %such that the learnt model dynamics $\hat f_\vtheta$ in (\ref{eqn:additive_model}) can serve as a good model for the true system dynamics when evaluating a controller $\pi$?
Second, how do we formulate the \textit{forward} and \textit{backward reachable tube} problems over the NN modeled system dynamics?

%In this section, we present the algorithm used to tackle Problems \ref{prob1_safety_verification} and \ref{prob2_safe_states}.
%
%Before delving into the details, we highlight the different aspects of this problem.
%
%Firstly, a conservative bound is estimated for the modeling error.
%
%Secondly, the forward and backward reachable tube formulation is developed for NN modeled dynamics.
%
%Finally, for an unsafe policy, the subset of safe initial states is determined.
%
\subsection{Model-Based Reinforcement Learning}
In this section, we focus on the necessary requirements for the modeled dynamics $\hat f_\vtheta$ and discuss the estimation of the modeling error set $\sD$.
A summary of the model-based RL framework is presented in Appendix \ref{sec:modelRL}.
Recall that the learnt model $\hat{f}_\vtheta$ is represented using an NN  %, which can have $H$ number of hidden layers. 
%
%Then, $\hat f_\vtheta$ is parameterized using $\vtheta = \{ \vtheta^{(1)}, \vtheta^{(2)}, \hdots, \vtheta^{(H)}\}$
%
%such that the output from a layer $h \in \{1, 2, \hdots, H\}$ is given as
% \begin{equation}
    %$\vy^{(h)} = g_h(\vtheta^{(h)} \vy^{(h-1)}),$
    % \label{eqn:feedforward}
% \end{equation}
% where $\mW^{(h)}$ and $\vb^{(h)}$ denote the weight matrix and bias vector of layer $h$ and 
%where $g_h$ denotes an activation function.
%
and predicts the change in states $\Delta \hat{\vs}_{t+1} \in \sR^n$.
%
% The model $\hat f$ is trained using the observed state transitions $D = \{ ( \vs(t), \va(t), \vs(t+1))^{(i)} \}_{i=1}^{N}$.
%
To learn $\hat f_\theta$, an observed data set $D= \{ ( \vs_t, \va_t, \Delta \vs_{t+1})^{(i)} \}_{i=1}^{N}$ is first split into 
% an $80-20$ proportion to form a
a training data set $D_t$ and a validation data set $D_v$.
A supervised learning technique is then used to train $\hat f_\theta$ over $D_t$ by minimizing the prediction error
% \begin{equation}
  $ E = \frac{1}{N_t} \sum_{i=1}^{N_t} || \Delta \hat{\vs}_{t+1}^{(i)} - \Delta \vs_{t+1}^{(i)} ||^2,$
% \end{equation}
where $\Delta \hat{\vs}_{t+1}$ is the change predicted by the learnt model, $\Delta \vs_{t+1}$ is the true observed change in state and $N_t = |D_t|$.
With this notation, we now formalize the following necessary assumption for this work.
This assumption is required to ensure boundedness during analysis and is easily satisfied by NNs that use common activation functions like $\texttt{tanh}$ or $\texttt{sigmoid}$ \citep{usama2018towards}.

% $\theta^{(j)}, \; j = \{1, 2, \hdots, H\}$.
%
% Therefore, we modify the notation of the learnt model to $\hat{f}_{\mW, \vb}(\vs, \va)$ in order to specify the parameters $\mW = \{ \mW^{(1)}, \mW^{(2)}, \hdots, \mW^{(H)}\}$ and $\vb = \{ \vb^{(1)}, \vb^{(2)}, \hdots, \vb^{(H)}\}$.
%

%
\begin{assumption} $\hat{f}_{\vtheta}$ is Lipschitz continuous and $\forall j \in \{1, ..., n\}, \Delta \hat{\vs}_j \in
[-c, c]$, where $|c| < \infty$.
  \label{ass:boundedness}
\end{assumption}
%
% \begin{assumption} \label{ass:continuity}
% $\hat{f}_{\vtheta}$ is uniformly continuous and Lipschitz continuous.
% \end{assumption}
%
% Assumptions \ref{ass:boundedness} and \ref{ass:continuity} define the class of NNs which can be used to represent $\hat{f}_{\vtheta}$ in this work.
%
% https://vanderbei.princeton.edu/tex/unif_cont/uc3.pdfto bound the output between a range $[a, b]$ or $[0, c]$ respectively.
% %
% \begin{equation}
%     \hat{s}_j = \begin{cases} a + \frac{(b-a)}{2} (y_j^{(H)} - 1)
%     &\mbox{if } g_H := tanh() \\
%     c y_j^{(H)} &\mbox{if } g_H := sigmoid() \end{cases}, \; \forall j = 1, \hdots n.
% \end{equation}
%
% Similarly, Assumption \ref{ass:continuity}
% Lipschitz smoothness has also been shown to hold for NNs with commonly used activation functions
% and is required in our analysis to ensure that small perturbation to the inputs for $\hat f_\vtheta$  does not lead to significantly different outcomes. 
% , with works approximating the Lipschitz constant for NNs like $tanh$, $sigmoid$, $ReLU$, $linear$, $softmax$, etc.

\textbf{Modeling error:}
As mentioned earlier, the accuracy of the learnt model $\hat{f}_{\vtheta}$ depends on the quality of data and the NN being used, thereby resulting in some modeling error $\vd$ in the prediction of the next state.
Estimating modeling errors is an active area of research and is required for several existing works on safe RL \citep{akametalu2014reachability, gillula2012guaranteed}, and is complementary to our goal.
Since the primary contribution of this work is the development of a reachable tube formulation for model-based controllers that use NNs, we rely on existing techniques \citep{moldovan2015optimism} to estimate a conservative modeling error bound.
%
% Since we don't have true transitions for every possible state, we cannot compute the exact value of $\vd$.
%
% To address this issue,
We leverage the error estimates $\hat \vd_j = \hat \Delta s_{t+1}^{(j)} - \Delta s^{(j)}_{t+1},$ of $\hat f_\theta$, for the transition tuples in the validation set $D_v$ to construct the upper confidence bound $\vd^+ = [d_1, \; d_2, \hdots d_n]$ and lower confidence bound $\vd^- = [-d_1, -d_2, \hdots -d_n]$  for $\vd$, for each state dimension.
Let a high-confidence bounded error set $\sD$ be defined such that $\sD = \{\vd: \forall i \in \{1,...,n\}, \,\, \vd^-_i < \vd_i < {\vd}^+_i\}$.
%
% on the validation set $D_v$ to obtain an estimate of $\vd$.
%
% One possible way would be to take the 
%
% This prediction error is termed as the modeling error and it is estimated over the validation data set $D_v$.
%
% By analyzing the trend of the prediction error made by $\hat{f}_{\mW, \vb}$ while executing the feedforward pass in (\ref{eqn:feedforward}) on $D_v$, we estimate a $3-\sigma$ bound on the modeling error for each of the $n$ states.
%
% If $d_i^\sigma$ denotes the $3-\sigma$ bound for each state, then a bounded error set $\sD$ is defined such that $\forall \vd \in \sD, \; \underline{\vd} < \vd < \overline{\vd}$, where $\underline{\vd} = [-d_1^\sigma, \; -d_2^\sigma, \hdots -d_n^\sigma]$ represents the lower bound and $\overline{\vd} = [d_1^\sigma, \; d_2^\sigma, \hdots d_n^\sigma]$ represents the upper bound of the modelling error.
%
We then use $\sD$ to represent the augmented learnt system dynamics as
\begin{equation}
    \hat{f}^{(r)}_{\vtheta} := \hat{f}_{\vtheta}(\vs, \va) + \vd, \; \;  \vd \in \sD.
    % \; \vd \in \sD. 
    \label{eqn:aug_r_dynamics}
\end{equation}
%
% \begin{assumption}
% The control inputs $\va$ and $\vd$ are drawn from closed and bounded sets, i.e., finite sets $\mathcal{A} \subset \mathbb{R}^m$ and $\mathcal{D} \subset \mathbb{R}^m$ respectively.
% \end{assumption}

% It is worth noting that our work shares the same limitation as other works on model-based RL, i.e., the verification framework depends on the quality of the training data and the flexibility offered by the NN model  \citep{akametalu2014reachability, gillula2012guaranteed}.
% %
% Since the primary contribution of this work is the development of a reachable set formulation for model based controllers that use NNs, we rely on existing techniques to estimate the modeling error.
% %
% Thus, a high-confidence bound is used to conservatively estimate the error, as is observed in the work \citet{moldovan2015optimism}.
% %
% By using a conservative error bound, the chances of falsely labeling an unsafe state as safe is reduced, and in this work we empirically validate this claim on the simulated problems.

\subsection{Reachable Tube Formulation}

For an exhaustive verification technique on a continuous state and action space, it is infeasible to sample trajectories from \textit{every} point in the given initial state and further, to verify whether \textit{all} these trajectories satisfy the safety constraints.
Therefore, reachable sets are usually (approximately) represented as convex polyhedrons and their evolution is tracked by pushing the boundaries of this polyhedron according to the system dynamics.
However, as convex polyhedrons can lead to large approximation errors, we leverage the level set method \citep{mitchell2005time} to compute the boundaries of the reachable tube for NN-based models and controllers at every time instant.
With the level set method, even non-convex boundaries of the reachable tube can be represented, thereby ensuring an accurate computation of the reachable tube \citep{mitchell2005time}.
Since the non-linear, non-convex structure of NNs is not suitable for analytical verification techniques, the reachability analysis gives an efficient, simulation-based approach to analyze the safety of the controller.
Therefore, in this subsection, we formulate the reachable tube for a NN modeled system dynamics $\hat{f}_{\vtheta}$, but first we formally define the forward reachable tube and backward reachable tube for a policy $\pi(\vs)$.

\begin{figure}[t]
\begin{center}
%\includegraphics[width=8cm]{images/Flowchart_MBRL.png}
%\framebox[4.0in]{$\;$}
%\fbox{\rule[-.5cm]{0cm}{4cm} \rule[-.5cm]{4cm}{0cm}
%\includegraphics[width=9cm]{images/Flowchart_MBRL.png}}
\includegraphics[width=0.8\textwidth]{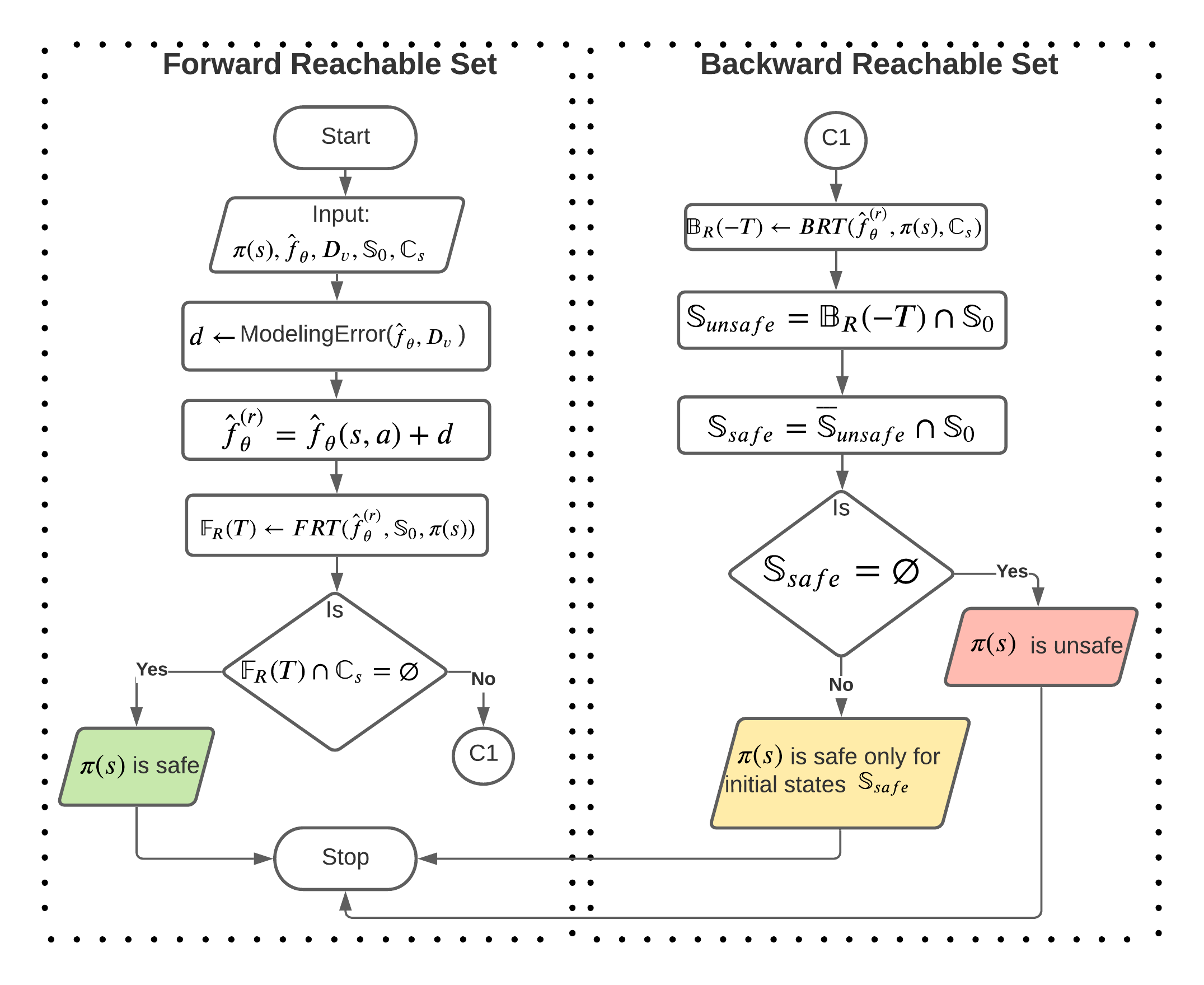}
\end{center}
\caption{Flowchart of the proposed safety verification algorithm. The rectangle on the left represents the flow for computing the forward reachable tube (FRT), which can only state if $\pi(\vs)$ is safe or unsafe for $\sS_0$. The rectangle on the right presents the flow for computing the backward reachable tube (BRT), which is invoked if $\pi(\vs)$ is unsafe. The BRT analysis can compute the subset of safe initial states $\sS_{safe}$ for $\pi(\vs)$, if such a set exists.}
\label{fig:flowchart}
\end{figure}

\textbf{Forward reachable tube (FRT):}
It is the set of all states that can be \textit{reached from} an initial set $\sS_0$, when the trajectories $\xi$ are executed under policy $\pi(\vs)$ and system dynamics $\hat{f}^{(r)}_{\vtheta}(\vs, \va, \vd)$.
The FRT is computed over a finite length of time $T$ and is formally defined as
\begin{equation}
    \!\!\!\sF_R(T) := \{ \vs : \forall \vd \in \sD, \vs(\cdot) \text{ satisfies } \dot{\vs} = \hat{f}^{(r)}_{\vtheta}(\vs, \va, \vd), \text{ where } \va = \pi(\vs), \; \vs_{t_0} \in \sS_0, t_f = T \},
    \label{eqn:FRT_def}
\end{equation}
where $t_0$ and $t_f$ denote the initial and final time of the trajectory $\xi$, respectively.

\textbf{Backward reachable tube (BRT):}
It is the set of all states which \textit{can reach} a given bounded target set $\sT \subset \sR^n$, when the trajectories $\xi$ are executed under policy $\pi(\vs)$ and system dynamics $\hat{f}^{(r)}_{\vtheta}(\vs, \va, \vd)$.
The BRT is also computed for a finite length of time, with the trajectories starting at time $t_0 = -T$ and ending at time $t_f = 0$.
It is denoted as
\begin{equation}
\begin{split}
\sB_R(-T) := \{ \vs_0 : \forall \vd \in \sD, \vs(\cdot) &\text{ satisfies } \dot{\vs} = \hat{f}^{(r)}_{\vtheta} (\vs, \va, \vd), \text{ where } \va = \pi(\vs)\\ &\text{ with } \vs_{t_0} = \vs_{-T}; \; \vs_{t_f} \in \sT, t_f \in [-T, 0] \}.
\end{split}
\end{equation}
The key difference between the FRT and BRT is that, for the former, the initial set of states are known, whereas for the latter, the final set of states are known.

\paragraph{Outline:}
The flowchart of the safety verification framework proposed in this work is presented in Fig. \ref{fig:flowchart}.
Given a model-based policy $\pi(\vs)$, the set of initial states $\sS_0$ and the set of constrained states $\sC_s$, the first step is to estimate the bounded set of modeling error $\sD$, as discussed in Section 3.1.
%
%Since the true system dynamics is unavailable, the estimated modeling error is used to develop the augmented learnt system dynamics $\hat{f}^{(r)}_\vtheta$, which serves as a close substitute for the true system dynamics.
%
Using $\hat{f}^{(r)}_\vtheta$, the FRT is constructed from the initial set $\sS_0$ and it contains all the states reachable by $\pi(\vs)$ over a finite time $T$.
Thus, if the FRT contains any state from the unsafe region $\sC_s$, $\pi(\vs)$ is deemed \textit{unsafe}.
Therefore, the solution to Problem \ref{prob1_safety_verification}, is determined by analyzing the set of intersection of the FRT with $\sC_s$ as
%\begin{equation}
%    \sF_R(T) \cap \sC_s = \begin{cases} = \emptyset
%    &\mbox{if } \pi \text{ is safe} \\
%    \neq \emptyset &\mbox{if } \pi \text{ is not safe.} \end{cases}
%\end{equation}
%
\begin{equation}
    \pi = \begin{cases} \text{ safe} 
    &\mbox{if } \,\, \sF_R(T) \cap \sC_s = \emptyset, \\
    \text{ unsafe} &\mbox{if }\,\, \sF_R(T) \cap \sC_s \neq \emptyset.  \end{cases}
\end{equation}
If $\pi(\vs)$ is classified as \textit{safe} for the entire set $\sS_0$, then no further analysis is required.
However, if $\pi(\vs)$ is classified as \textit{unsafe}, we proceed to compute the subset  $\sS_\text{safe} \subset \sS_0$ of initial states for which $\pi(\vs)$ generates safe trajectories.
$\sS_{safe}$ is the solution to Problem \ref{prob2_safe_states} and allows an unsafe policy to be deployed on a real system, with restrictions on the starting states.
To this end, the BRT is computed from the unsafe region $\sC_s$, to determine the set of trajectories (and states) which terminate in $\sC_s$.
The intersection of the BRT with $\sS_0$ determines the set of unsafe initial states $\sS_{unsafe}$.
To determine $\sS_{safe}$, we utilize the following properties, (a) $\sS_{safe} \cup \sS_{unsafe} = \sS_0$, and (b) $\sS_{safe} \cap \sS_{unsafe} = \emptyset$,
% \begin{equation}
%   %\begin{split}
%     \sS_{safe} \cup \sS_{unsafe} = \sS_0, \quad  \quad
%     \sS_{safe} \cap \sS_{unsafe} = \emptyset,
%     %\end{split}
%     \label{eqn:set_int}
% \end{equation}
and compute
 \begin{equation}
    \sS_{safe} = \overline{\sS}_{unsafe} \cap \sS_0.
\end{equation}
If $\sS_{safe} \neq \emptyset$, then we have identified the safe initial states for $\pi(\vs)$, otherwise, it is concluded that there are no initial states in $\sS_0$ from which $\pi(\vs)$ can generate safe trajectories.
\paragraph{Mathematical Formulation:}
This section presents the mathematical formulation to compute the BRT.
The FRT can be computed with a slight modification to the BRT formulation and this is discussed in the end of this section.
%
%In the safety verification framework, the BRT will be generated for the constrained regions $\sC_s$, to first determine the unsafe initial states $\sS_{unsafe} \subset \sS_0$, followed by the computation of $\sS_{safe}$ for Problem \ref{prob2_safe_states}.
%

Recall, for the BRT problem, there exists a target set $\sT \subset \sR^n$ which the agent has to reach in finite time, i.e., the condition on the final state is given as $\vs_{t_f} \in \sT$.
Conventionally, for the BRT formulation, the final time $t_f = 0$ and the starting time $t_0 = -T$, where $0 < T < \infty$.
%
%Additionally, a policy $\pi(\vs)$ to be evaluated is already provided.
%
%$\pi(\vs)$ is trained to reach $\sT$, while optimizing a cost function $J$.
%
%To develop an optimal policy $\pi(\vs)$ to reach $\sT$, we first need to define the cost function $J$.
%
When evaluating a policy $\pi(\vs)$, the controller input is computed by the given policy as $\va = \pi(\vs)$.
However, following the system dynamics $\hat{f}_\vtheta^{(r)}$ in (\ref{eqn:aug_r_dynamics}), the modeling error $\vd$ is now included in the system as an adversarial input, whose value at each state is determined so as to maximize the controller's cost function.
We use the HJ PDE to formulate the effect of the modeling error on the system for computing the BRT, but first we briefly review the formulation of the HJ PDE with an NN modeled system dynamics $\hat{f}_\vtheta$ in the following.

For an optimal controller, we first define the cost function which the controller has to minimize.
Let $C(\vs_t, \va_t)$ denote the running cost of the agent, which is dependent on the state and action taken at time $t \in [-T, 0]$.
Let $g(\vs_{t_f})$ denote the cost at the final state $\vs_{t_f}$.
%
%Then, to evaluate the given policy $\pi(\vs)$, the following optimization problem is framed
%
Then, the goal of the optimal controller is to find a series of optimal actions such that
\begin{equation}
\begin{split}
    %\min_{\va(\cdot)} 
    \min_{\va_{\tau}(\cdot)} &\left(\int_{-T}^{0} C(\vs_{\tau}, \va_{\tau}) d \tau + g(\vs_{t_f}) \right)\\
    \text{subject to  } \dot{\vs} &= \hat{f}_{\vtheta}(\vs, \va), \,\,\,
    %\va &= \pi(\vs), \\
    \vs_{t_f} \in \sT,
\end{split}
\label{eqn:J_v1}
\end{equation}
where $\va_{\tau} \in \sA$ and $\hat{f}_\vtheta$ is the NN modeled system dynamics.
%
%Therefore, to consider the worst case scenario, the modeling error is formulated as an adversarial input which is trying to maximize $J$, while the policy input is trying to minimize $J$.
%
The above optimization problem is solved using the \textit{dynamic programming} approach \citep{smith1991dynamic}, which is based on the \textit{Principle of Optimality} \citep{troutman2012variational}.
Let $V(\vs_t, t)$ denote the value function of a state $\vs$ at time $t \in [-T, 0]$, such that
\begin{equation}
    \!\!\! V(\vs_t, t) = %\min_{\va(\cdot)} 
    \min_{\va_\tau(\cdot)} \left[ \int_{t}^{0} C(\vs_{\tau}, \va_{\tau}) d \tau + g(\vs_{t_f}) \right]  = \min_{\va_\tau(\cdot)} \left[ \int_{t}^{t + \delta} \!\!\! C(\vs_{\tau}, \va_{\tau}) d \tau + V(\vs_{t+\delta}, t + \delta ) \right],
    \label{eqn:V_def}
\end{equation}
where $\delta > 0$.
$V(\vs_t, t)$ is a quantitative measure of being at a state $\vs$, described in terms of the cost required to reach the goal state from $\vs$.
%
% Since, the optimization problem in (\ref{eqn:J_v1}) seeks to compute a sequential optimal solution, using the \textit{Principle of Optimality} \citep{troutman2012variational}, the value function can be iteratively computed as
% \begin{equation}
    % V(\vs_t, t) = \min_{\va_\tau(\cdot)} \int_{t}^{t + \delta} C(\vs_{\tau}, \va_{\tau}) d \tau + V(\vs_{t+\delta}, t + \delta )
% \label{eqn:rec_value}
% \end{equation}
%
Then, using the Taylor series expansion, $V(\vs_{t + \delta}, t + \delta)$ is approximated around $V(\vs_t, t)$ in (\ref{eqn:V_def}) to derive the HJ PDE as
\begin{equation}
\begin{split}
\frac{dV}{dt} + \min_{\va} \left[ \nabla V \cdot \hat{f}_{\vtheta}(\vs, \va) + C(\vs, \va) \right] &= 0, \\
V(\vs_{t_f}, t_f) &= g(\vs_{t_f}),
\end{split}
\label{eqn:HJI_PDE_value}
\end{equation}
where $\nabla V \in \sR^n$ is the spatial derivative of $V$.
Additionally, the time index has been dropped above and the dynamics constraint in (\ref{eqn:J_v1}) has been included in the PDE.
Equation (\ref{eqn:HJI_PDE_value}) is a terminal value PDE, and by solving (\ref{eqn:HJI_PDE_value}), we can compute the value of a state $V(\vs_t, t)$ at any time $t$.

We now discuss how the formulation in (\ref{eqn:HJI_PDE_value}) can be modified to obtain the BRT.
It is noted that along with computing the value function, the formulation in (\ref{eqn:HJI_PDE_value}) also computes the optimal action $\va$.
However, in Problems \ref{prob1_safety_verification} and \ref{prob2_safe_states}, the optimal policy $\pi(\vs)$ is already provided.
Therefore, the constraint $\va = \pi(\vs)$ should be included in problem (\ref{eqn:J_v1}), thereby avoiding the need of minimizing over actions $\va \in \sA$ in (\ref{eqn:HJI_PDE_value}).
Additionally, as discussed in Section 3.1, the NN modeled system dynamics $\hat{f}_\vtheta$ may not be a good approximation of the true system dynamics $f$.
Instead, the augmented learnt system dynamics $\hat{f}^{(r)}_\vtheta$ in (\ref{eqn:aug_r_dynamics}) is used in place of $\hat{f}_\vtheta$ in (\ref{eqn:HJI_PDE_value}), since it better models the true dynamics at a given state.
However, by including $\hat{f}_\vtheta^{(r)}$ in (\ref{eqn:HJI_PDE_value}), the modeling error $\vd$ is now included in the formulation.
The modeling error $\vd \in \sD$ is treated as an adversarial input which is trying to drive the system away from it's goal state by taking a value which maximizes the cost function at each state.
Thus, to account for this adversarial input, the formulation in (\ref{eqn:HJI_PDE_value}) is now maximized over $\vd$.
%
%Thus, at any state $\vs$ the modeling error is computed as the vector which maximizes the cost at that state.

Lastly, the BRT problem is posed for a set of states and not an individual state.
Hence, an efficient representation of the target set is required to propagate an entire set of trajectories at a time, as opposed to propagating individual trajectories.
\begin{assumption} \label{ass:level_set}
The target set $\sT \subset \sR^n$ is closed and can be represented as the zero sublevel set of a bounded and Lipschitz continuous function $l: \sR^n \
\rightarrow \sR$, such that, $\sT = \{ \vs : l(\vs) \le 0\}$.
\end{assumption}
The above assumption defines a function $l$ to check whether a state lies inside or outside the target set.
If $\sT$ is represented using a regular, well-defined geometric shape (like a sphere, rectangle, cylinder, etc.), then deriving the level set function $l(\vs)$ is straight forward, whereas, an irregularly shaped $\sT$ can be represented as a union of several well-defined geometric shapes to derive $l(\vs)$.

For the BRT problem, the goal is to determine all the states which can reach $\sT$ within a finite time.
The path taken by the controller is irrelevant and only the value of the final state is used to determine if any state $\vs \in \sB_R(-T)$.
From Assumption \ref{ass:level_set}, the terminal condition $\vs_{t_f} \in \sT$ can be restated as $l(\vs_{t_f}) \le 0$.
Thus, to prevent the system from reaching $\sT$, the adversarial input $\vd$ tries to maximize $l(\vs_{t_f})$, thereby pushing $\vs_{t_f}$ as far away from $\sT$ as possible.
Therefore, the cost function in (\ref{eqn:J_v1}) is modified to $J = l(\vs_{t_f})$.
Additionally, any state which can reach $\sT$ \textit{within} a finite time interval $T$ is included in the BRT.
Therefore, if any trajectory reaches $\sT$ at some $t_f<0$, it shouldn't be allowed to leave the set.
Keeping this in mind, the BRT optimization problem can be posed as
\begin{equation}
\begin{split}
\max_{\vd(\cdot)} & \left( \min_{t \in [-T, 0]} l(\vs_{t_f}) \right)\\
\text{subject to: } \dot{\vs} &= \hat{f}^{(r)}_{\vtheta} (\vs, \va, \vd), \,\,\,
\va = \pi(\vs), \,\,\,\,
l(\vs_{t_f}) \le 0,
\end{split}
\label{eqn:J_level}
\end{equation}
where the inner minimization over time prevents the trajectory from leaving the target set.
Then, the value function for the above problem is defined as
\begin{equation}
    V_{R}(\vs_t, t) := \max_{\vd(\cdot)} l(\vs(t_f)).
\label{eqn:v_r}
\end{equation}
Comparing this with (\ref{eqn:V_def}), it is observed that the value of a state $\vs$ is no longer dependent on the running cost $C(\vs, \va)$.
This doesn't imply that the generated trajectories are not optimal w.r.t. action $\va$, because the running cost is equivalent to the negative reward function, for which $\pi(\vs)$ is already optimized.
Instead, $V_R$ solely depends on whether the final state of the trajectory lies within the target set or not, i.e., whether or not $\vs_{t_f} \in \sT$.
Thus, the value function $V_R$ for any state $\vs$ is equal to $l(\vs_{t_f})$, where $\vs_{t_f}$ is the final state of the trajectory originating at $\vs$.
%
%This doesn't imply that the generated trajectories are not optimal w.r.t. action $\va$ because the running cost is equivalent to the negative reward function for which $\pi(\vs)$ is optimized.
%
Then, the HJ PDE for the problem in (\ref{eqn:J_level}) is stated as
\begin{equation}
\begin{split}
\frac{dV_R}{dt} + min \{ 0, H^*(\vs, \nabla V_R(\vs_t, t), t) \} &= 0, \\
V_R(\vs_{t_f}, t_f) &= l(\vs_{t_f}), \\
\text{where } H^* &= \max_{\vd} \left( \nabla V_R \cdot \hat{f}^{(r)}_{\vtheta}(\vs, \pi(\vs), \vd) \right), \\
\end{split}
\label{eqn:BRT_v1}
\end{equation}
where $H^*$ represents the optimal Hamiltonian.
Since we are computing the BRT, $min\{ 0, H^*(\vs, \nabla V_R(\vs_t, t), t) \}$ in the PDE above ensures that the tube grows only in the backward direction, thereby preventing a trajectory which has reached $\sT$ from leaving.
In Problem (\ref{eqn:BRT_v1}), the optimal action has been substituted by $\pi(\vs)$ and the augmented learnt dynamics is used instead of $\hat{f}_\vtheta$.
The Hamiltonian optimization problem can be further simplified to derive an analytical solution for the modeling error.
By substituting the augmented dynamics from (\ref{eqn:aug_r_dynamics}), the optimization problem can be re-written as
\begin{equation}
    H^* = \nabla V_R \cdot \hat{f}_{\vtheta}(\vs, \va) + \max_{\vd} \nabla V_R \cdot \vd.
\end{equation}
Expanding $\nabla V_R = [p_1, p_2, \hdots, p_n]^T \in \sR^n$, the vector product $\nabla V_R \cdot \vd = p_1 d_1 + p_2 d_2 + \hdots + p_n d_n$.
Therefore, to maximize $\nabla V_R \cdot \vd$, the disturbance control is chosen as
\begin{equation}
d_i = \begin{cases} d_i
    &\mbox{if } p_i > 0 \\
      -d_i
    &\mbox{if } p_i < 0 \end{cases} , \forall i=1,\hdots n.
\label{eqn:H_opt2}
\end{equation}
With this analytical solution, the final PDE representing the BRT is stated as
\begin{equation}
\begin{split}
\frac{dV_R}{dt} + min \{ 0, H^*(\vs, \nabla V_R(\vs_{t}, t), t) \} &= 0, \\
V_R(\vs_{t_f}, t_f) &= l(\vs_{t_f}), \\
\text{where } H^* = \nabla V_R \cdot \hat{f}_{\vtheta}(\vs, \va) + &(|p_1| d_1 + |p_2| d_2 + \hdots + |p_n| d_n).
\end{split}
\label{eqn:BRT_final}
\end{equation}
%
%Let the BRT constructed from the constrained regions $\sC_s$ be denoted as $\sB_C(-T)$.
%
%Then, the set of unsafe initial sets is determined as $\sS_{unsafe} = \sB_C(-T) \cap \sS_0$.
%
%The following statements 
%\begin{equation}
  %\begin{split}
%    \mathcal{S}_{safe} \cup \mathcal{S}_{unsafe} = \mathcal{S}_0, \quad \quad
%    \mathcal{S}_{safe} \cap \mathcal{S}_{unsafe} = \emptyset
    %\end{split}
%    \label{eqn:set_int}
%  \end{equation}
%are used to determine the safe initial set as
%  \begin{equation}
%    \mathcal{S}_{safe} = \overline{\mathcal{S}}_{unsafe} \cap \mathcal{S}_0.
%  \end{equation}

The value function $V_R(\vs_t, t)$ in (\ref{eqn:BRT_final}) represents the evolution of the target level set function backwards in time.
By finding the solution to $V_R$ in the above PDE, the level set function is determined at any time instant $t \in [-T, 0]$, thereby determining the BRT.
%
%However, finding the exact solution to (\ref{eqn:BRT_final}) is difficult.
%
From the result of Theorem 2 in \citet{mitchell2005time}, it is proved that the solution of $V_R$ in (\ref{eqn:BRT_final}) at any time $t$ gives the zero sublevel set for the BRT.
%%Thus, all $\vs$ for which $V_R(\vs_t, t) \le 0, t \in [-T, 0]$ belong to the BRT of $\sT$. 
Thus,
\begin{equation}
    \sB_R(-T) = \{ \vs : V_R(\vs_t, t) \le 0, \; t \in [-T, 0] \}.
\end{equation}
The solution to $V_R$ can be computed numerically by
using existing solvers for the level set method.
A brief note on the implementation of the algorithm is included in subsection A.3 in the Appendix.

There are a few things to note about the formulation in (\ref{eqn:BRT_final}).
First, Equation (\ref{eqn:BRT_final}) assumes that $\sT$ is a desired goal state.
However, the formulation can be modified if $\sT$ is an unsafe set, in which case, the adversarial modeling error tries to minimize the Hamiltonian.
Similarly, the input $\vd$ can represent any other disturbance in the system, either adversarial or cooperative.
Second, to compute the FRT, the formulation in (\ref{eqn:BRT_final}) is modified from a final value PDE to an initial value PDE.

%\subsection{Algorithm}

%% file: experiments.tex
\section{Experiments}
\begin{figure}[t]
\begin{center}
\includegraphics[width=0.32\textwidth]{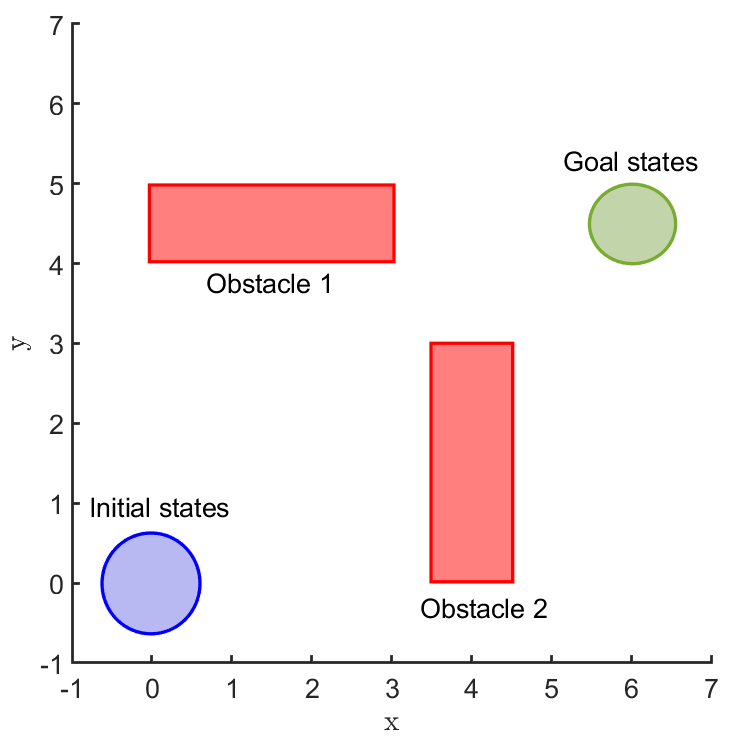}
\includegraphics[width=0.32\textwidth]{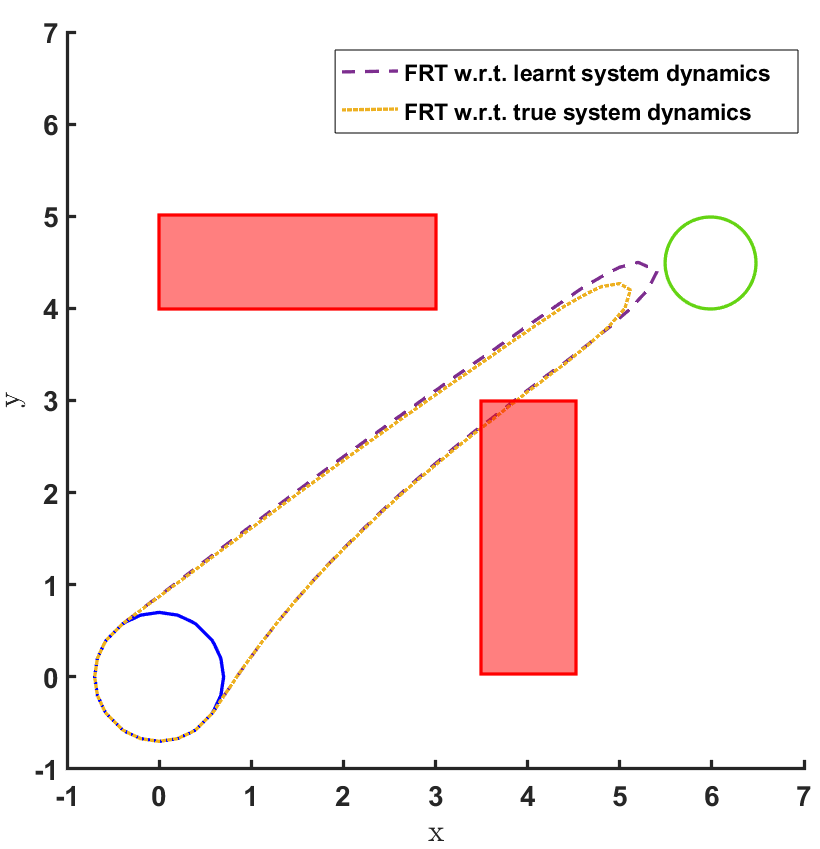}
\includegraphics[width=0.32\textwidth]{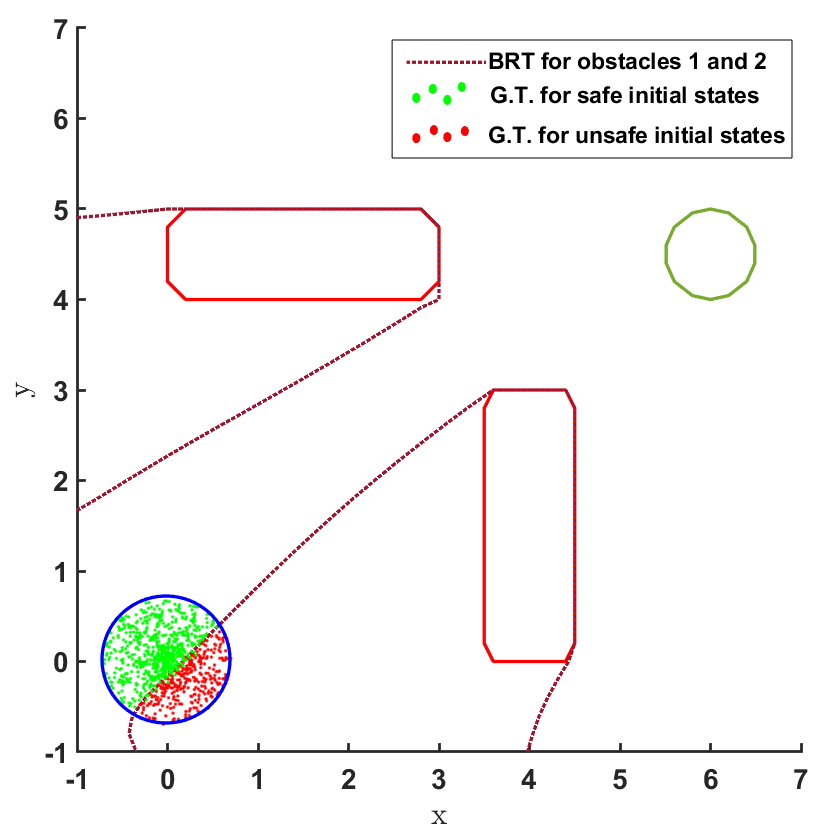}
\end{center}
\caption{(Left) The environment for the safe land navigation problem. (Middle) The FRTs for both the augmented learnt model dynamics and true system dynamics classify the controller as unsafe. (Right) The BRT computed from obstacles 1 and 2 for the given controller. $\sS_{safe}$ computed by the proposed BRT algorithm is compared with the ground truth (G.T.) data for the safe initial states, marked in green.}
\label{fig:2D_init}
\end{figure}

In our experiments, we aim to answer the following questions:  \textbf{(a)} Can safety verification be done for an NN-based $\pi$ and $\hat f_\theta$ using FRT?, and \textbf{(b)} Can $\sS_\text{safe}$ be identified using BRT if $\pi$ is deemed unsafe?
To answer these two questions, we demonstrate results on the following  domains inspired by real world safety-critical problems, where RL controllers developed using a learnt model can be appealing, as they can adapt to transition dynamics involving friction, air-drag, wind, etc., which might be hard to explicitly model otherwise. 

\noindent \textbf{Safe land navigation: } Navigation of a ground robot in an indoor environment is a common application which requires the satisfaction of safety constraints by $\pi$ to avoid collision with obstacles.
For this setting, we simulate a ground robot which has continuous states and actions.
%
% , i.e., the state vector $\vs = [x, y]^T$ representing the $x$ and $y$ coordinates of the car and the control input vector $\va = [v, \psi]$ comprises of the velocity $v$ and the direction $\psi$ of the vehicle.
%
The initial configuration of the domain is shown in Fig. \ref{fig:2D_init}.
The set of initial and goal states are represented by circles and the obstacles with rectangles.

\begin{figure}[t]
\begin{center}
\includegraphics[width=0.32\textwidth]{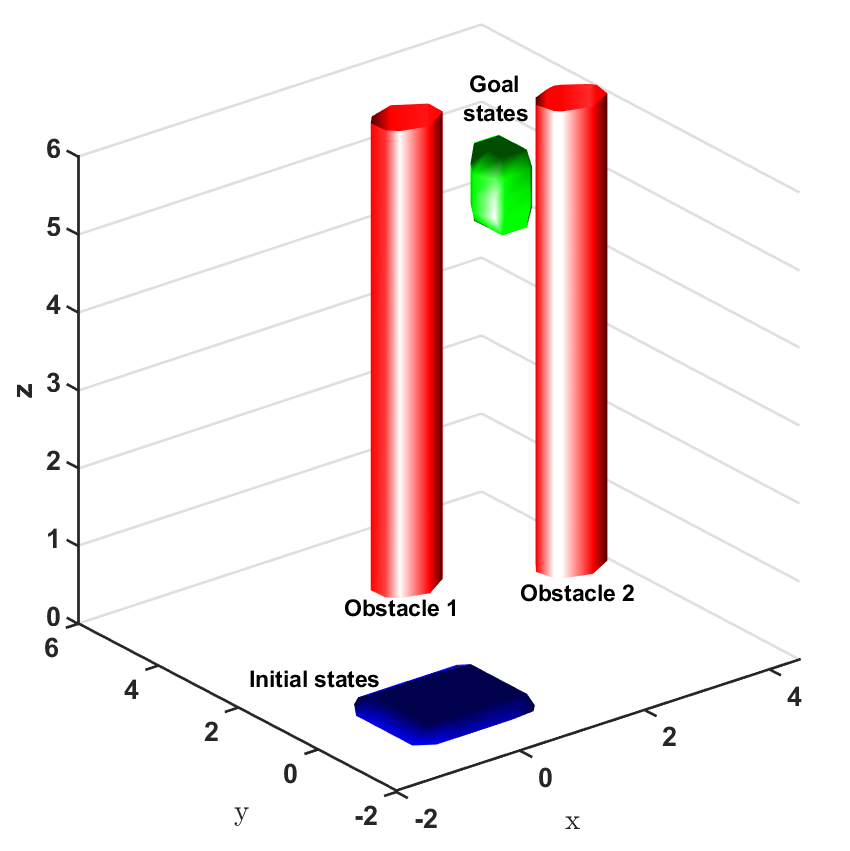}
\includegraphics[width=0.32\textwidth]{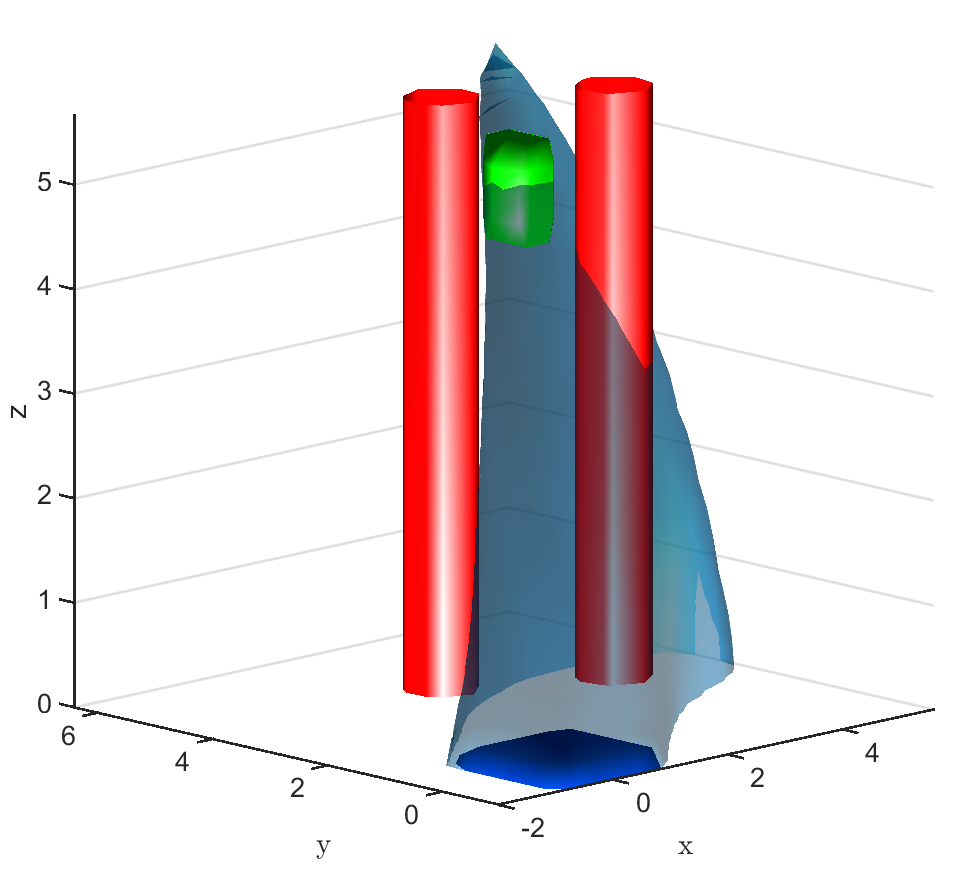}
\includegraphics[width=0.32\textwidth]{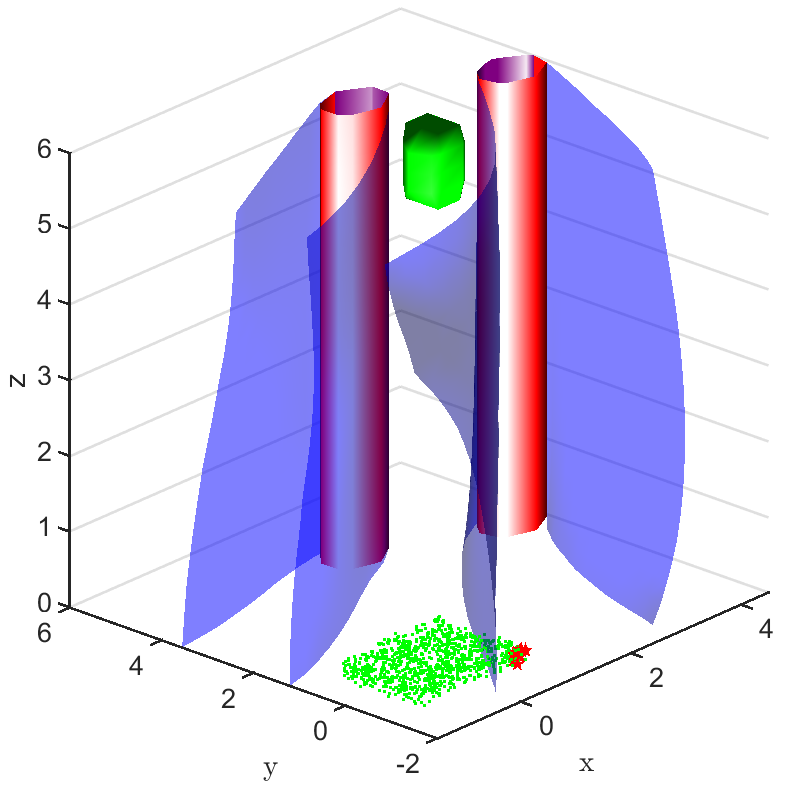}
\end{center}
\caption{(Left) The safe aerial navigation domain. (Middle) The FRT is computed for the augmented learnt dynamics. For clarity in 3D, FRT with true dynamics is not plotted; instead the approximation quality can be better visualized in BRT. (Right) Comparison of $\sS_{safe}$ computed by the proposed BRT algorithm with the ground truth data of the safe initial states, marked in green  (and unsafe states marked in red), it is observed that the BRT conservatively approximates  $\sS_{safe}$.}
\label{fig:3D_init}
\end{figure}

\noindent \textbf{Safe aerial navigation: }
This domain simulates a navigation problem in an urban environment for an unmanned aerial vehicle (UAV).
Constraints are incorporated while training $\pi$ to ensure that collision is avoided with potential obstacles in its path.
States and actions are both continuous and the initial configuration of the domain is shown in Fig \ref{fig:3D_init}.
The set of initial and goal states are represented using cuboids, and the obstacles with cylinders.

\paragraph{Analysis:} To address the questions with respect to the above mentioned domains, we first train a NN based $\hat f_\theta$ to estimate the dynamics using sampled transitions.
This $\hat f_\theta$ is then also used to learn a NN based controller $\pi$ which is trained with a cost function designed to mitigate collisions.
For brevity, only the representative results for this $\pi$ are discussed here; implementation details and more experimental results are available in Appendix \ref{sec:modelRL}, \ref{sec:implementation}, \ref{sec:results_land} and \ref{sec:results_aerial}.

To address the first question, the FRT is computed for both the domains over the augmented learnt dynamics $\hat{f}^{(r)}_\vtheta$ as shown in Fig. \ref{fig:2D_init} and Fig. \ref{fig:3D_init}.
Additionally, for land navigation we also compute the FRT over the true system dynamics $f$, which serves as a way to validate the safety verification result of $\pi$ from the proposed framework.
It is observed that for both the domains, FRTs deem the given policy $\pi$ as \textit{unsafe}, since the FRTs intersect with one of the obstacles.
Even when $\pi$ is learnt using a cost function designed to avoid collisions, the proposed safety verification framework successfully brings out the limitations of $\pi$, which may have resulted due to the use of function approximations, ill-specified hyper-parameters, convergence to a local optimum, etc.

For the second question, the BRT is computed from both the obstacles for the given controller $\pi$, as shown in Fig. \ref{fig:2D_init} and Fig. \ref{fig:3D_init}.
To estimate the accuracy of the BRT computation, we compare the computed $\sS_{unsafe}$ and $\sS_{safe}$ sets with the ground truth (G.T.) data generated using random samples of possible trajectories.
%
% The G.T. data was generated by sampling random trajectories from $\sS_0$ and identifying the initial states corresponding to unsafe trajectories (marked in red) and safe trajectories (marked in green).
%
It is observed that the BRT from obstacle 1 does not intersect with $\sS_0$, implying that all trajectories are safe w.r.t. obstacle 1.
%
% This result can also be cross-verified by observing the FRT computed, since $F_R(T) \cap \sC_s^{(1)} = \emptyset$.
%
However, the BRT from obstacle 2 intersects with $\sS_0$ and identifies the subset of initial states which are unsafe.
Such an information can be critical to safely deploy even an unsafe controller just by restricting its starting conditions. 
%
%
% It is observed that the proposed BRT algorithm computes $\sS_{safe}$ and $\sS_{unsafe}$ fairly accurately for this problem setting.
%
% Additional results for different models and controllers are presented in Appendix \ref{sec:results_land}.

%\subsection{Experiment 1:}

%% file: Appendix.tex
\section{Appendix}

\input{background_related_work}

\subsection{Algorithm for Model-Based RL}
\label{sec:modelRL}
RL follows the formal framework of a Markov decision process (MDP) which is represented using a tuple $M := (\sS, \sA, f, R)$ where $\sS$, $\sA$ and $f$ denote the state space, action space and true system dynamics, as defined in Section 2.
The last element in the MDP tuple is the reward function $R: \sS \times \sA \rightarrow \sR$ which quantifies the \textit{goodness} of taking an action $\va_t$ in state $\vs_t$, by computing the scalar value $r_t := R(\vs_t, \va_t)$.
Then the goal of the RL agent is to develop a policy (controller) $\pi$ which maximizes the expectation of the total return $G = \sum_{t=1}^T \gamma^{t-1} r_t$ over time $T$, where $\gamma \in [0, \; 1)$ is the discount factor.
In the model-based RL framework, the policy $\pi$ can simply be a planning algorithm or it can be a function approximator.

Under the model-based RL framework, the true dynamics function $f$ is unknown and needs to be learnt from observed data samples collected from the actual environment.
The learnt model is represented using a function approximator as $\hat{f}_\vtheta$.
A generic algorithm for model-based RL is provided in Algorithm 1.
To train the model, an initial training data set $D^{(0)}$ is first generated by sampling trajectories $\xi$ using a randomly initialized policy $\pi$.
In any iteration $k$, the data tuples $D^{(k)} = \{ (\vs_t, \va_t, \Delta \vs_{t+1})_i \}_{i=1}^{N_k}$ are used to train $\hat{f}_\vtheta$ by minimizing the prediction error $E$ %$E := \frac{1}{N} \sum_{i=1}^{N_k} || \hat{f}(s_t, a_t) - s_{t+1} ||^2$ 
using a  supervised learning technique.
This trained model $\hat{f}_\vtheta$ is then used to update the policy $\pi$ or used by a planning algorithm to generate new trajectories and improve the return $G$ obtained by the agent.
The data tuples obtained from the new trajectories are appended to the existing training data set to create $D^{(k+1)} = D^{(k)} \cup \{ \xi_i \}$ and the process continues until the performance of the controller converges.

\begin{algorithm}
\SetAlgoLined
\KwInput{$\pi^{(0)}$, $\hat{f}^{(0)}_\vtheta$}
 Generate trajectories $\xi$ using base policy $\pi^{(0)}$\;
 Collect and construct initial data set $D^{(0)} = \{ (\vs_t, \va_t, \Delta \vs_{t+1})_i \}_{i=1}^{N_0}$\;
 k = 0\;
 \While{$\Delta E > \epsilon$}{
  Learn dynamics model $\hat{f}^{(k)}_{\vtheta}(\vs, \va)$ over $D^{(k)}$ by minimizing error $E(D^{(k)})$\;
  \While{$\Delta G > \delta$}{
  At time $t$ plan through $\hat{f}^{(k)}_\vtheta(\vs,\va)$ to select action $\va_t$\;
  Execute action $\va_t$ and observe reward $r_t$ and next state $\vs_{t+1}$ (MPC)\;
  Obtain updated policy $\pi^{(k+1)}$ using backpropagation to maximize return $G = \sum_{t=1}^T\gamma^{t-1} r_t$\;
  Create data set $D^{(k+1)}$ by appending new observed data points\;
  }
  k = k+1\;
 }
 \caption{Generic model-based RL algorithm}
 \label{algo1}
\end{algorithm}

\subsection{Implementation of Proposed Safety Verification Framework}
\label{sec:implementation}
To implement the proposed safety verification framework in Fig. \ref{fig:flowchart}, the MATLAB helperOC \cite{helperOC_tool} developed and maintained by the HJ Reachability Group on Github was used.
This toolbox is dependent on the level set toolbox developed by \citet{LSM_toolbox}.
In the helperOC toolbox, the \textit{level set method} is used to provide a numerical solution to the PDE in (\ref{eqn:BRT_final}).
Level set methods have been used to track and model the evolution of dynamic surfaces (here the reachable tube) by representing the dynamic surface implicitly using a level set function at any instant of time \citep{osher1988fronts, osher2004level, sethian1999level}.
There are several advantages to using the level set algorithms, including the fact that they can represent non-linear, non-convex surfaces even if the surface merges or divides over time.
Additionally, it is easy to extend the theory to the higher dimensional state space.
However, the computational time increases exponentially with a dimension of the state space.
The helperOC \citet{helperOC_tool} has demonstrated results on problems with 10 dimensional space and since most physical systems can be represented within this limit, it is sufficient for the problems of interest in this work.

The helperOC \citet{helperOC_tool} comes with inbuilt numerical schemes to compute the Hamiltonian $H$, the spacial derivatives $\nabla V_R$ and the time derivative $\frac{dV_R}{dt}$ in (\ref{eqn:BRT_final}).
This toolbox is modified to solve the problem in (\ref{eqn:BRT_final}) by substituting the true system dynamics $f$ with the augmented learnt dynamics $\hat{f}^{(r)}_\vtheta(\vs, \va, \vd)$ and computing the action at any state $\vs$ as $\va = \pi(\vs)$.
Subsequently, to determine the level set function at any time instant, $\frac{dV}{dt}$ is computed using the Runge-Kutta integration scheme while the gradient of the level set function $\nabla V_R$ is computed using an upwind finite difference scheme.
Additionally, a Lax-Friedrichs approximation is used to stably evaluate the Hamiltoniaan \citep{Lax_fried}.

\subsection{Experiment Results: Safe Land Navigation}
\label{sec:results_land}
In the land navigation problem, a ground robot is moving in an indoor environment with obstacles.
The ground vehicle has the state vector $\vs = [x, y]^T$ representing the $x$ and $y$ coordinates of the car.
The control vector $\va = [v, \psi]$ comprises of the velocity $v$ and the direction $\psi$ of the vehicle.
The problem setting in shown in Fig. \ref{fig:2D_init}.
The set of initial and goal states are represented by circles centered at coordinates $(0.0, 0.0)$ and $(6.0, 4.5)$, respectively, and with radii of $0.7$ and $0.5$ units, respectively.
There are two rectangular obstacles in the environment with centers at $(1.5, 4.5)$ and $(4.0, 1.5)$.

To demonstrate the working of the proposed safety verification framework, different models and controllers were trained for the 2D land navigation problem setting.
Three different models were trained over training data sets of different sizes.
Model 1 was trained over 300 data samples, model 2 over 600 data samples, and model 3 over 1000 data samples.
To estimate the modeling error for all the three models, a $\pm 3\sigma$ bound was computed by first computing the standard deviation (s.d.) over the validation data sets (Table \ref{tab:sd_2d}).
\begin{table}
\begin{center}
\caption{Standard deviations (s.d.) computed for different models.} \label{tab:sd_2d}
\begin{tabular}{| c | c | c | c | }
  \hline
  & Model 1 & Model 2 & Model 3 \\ 
  \hline
  s.d. for state $x$ & 0.00114 & 0.00045 & 0.00031 \\  
 \hline
  s.d. for state $y$ & 0.00120 & 0.00053 & 0.00042 \\  
 \hline
\end{tabular}
\end{center}
\end{table}
Then, for each model, three different policies were sampled after training them for 100, 300 and 600 episodes, respectively.
We first present the FRT results for all controllers correspondong to every model, followed by the BRT results.

\begin{figure}[h]
\begin{center}
\includegraphics[width=0.32\textwidth]{images/FRT_300_100_v2.png}
\includegraphics[width=0.32\textwidth]{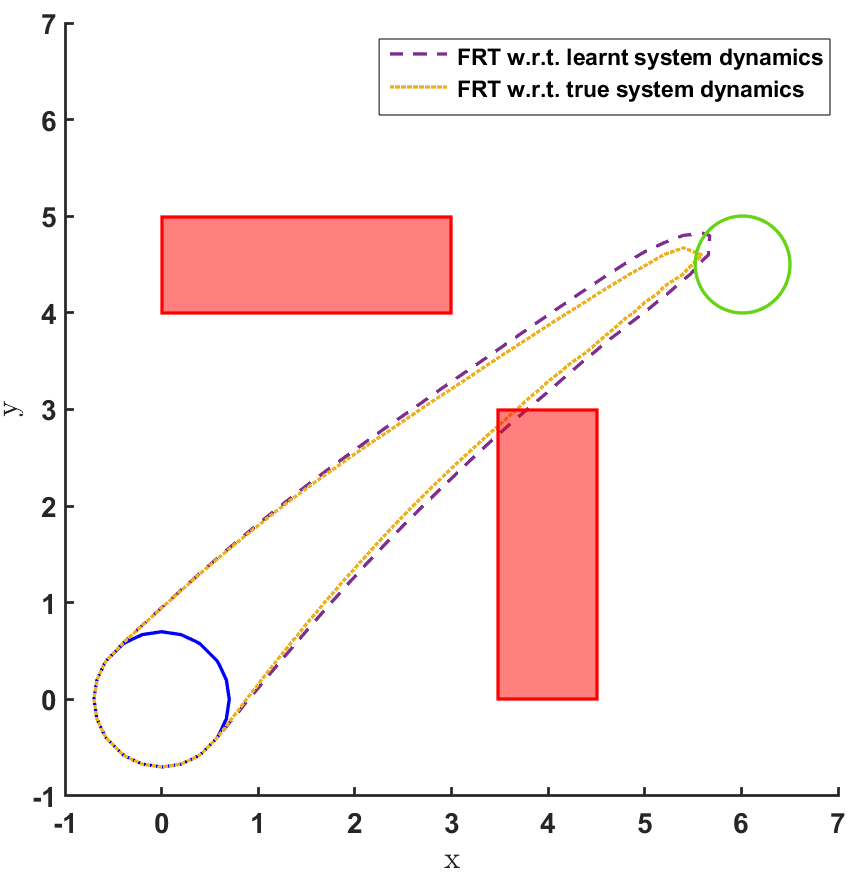}
\includegraphics[width=0.32\textwidth]{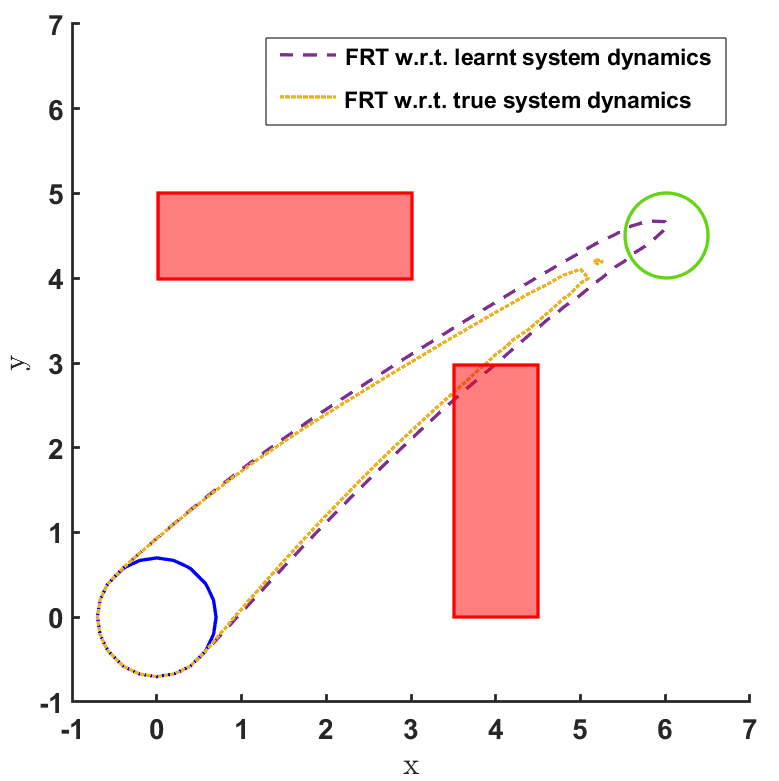}
\\
\includegraphics[width=0.32\textwidth]{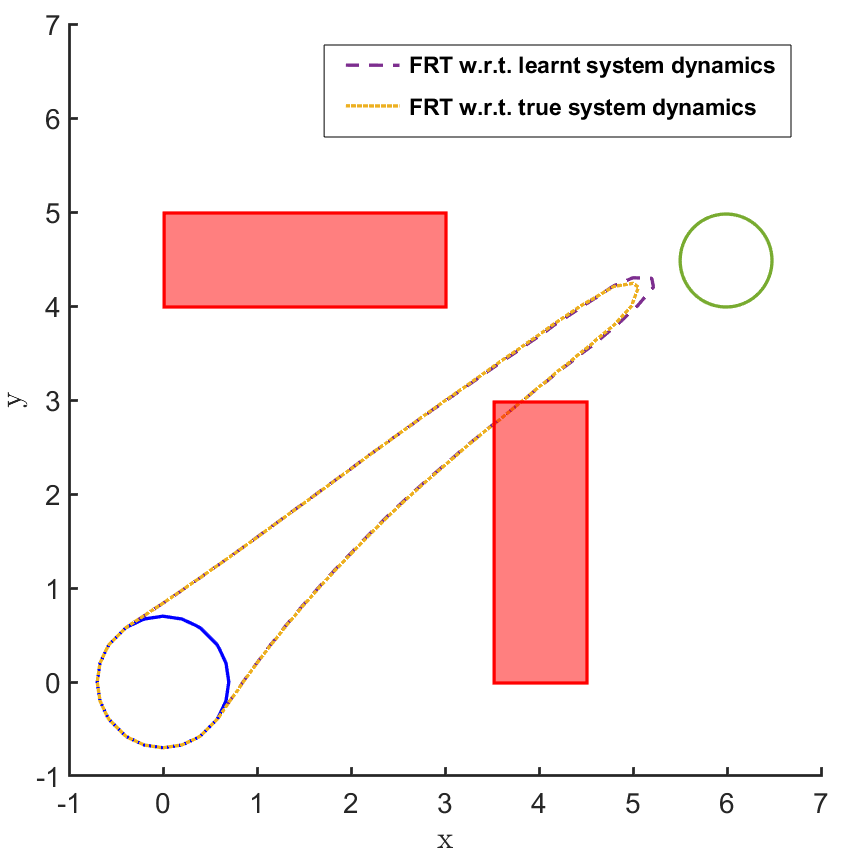}
\includegraphics[width=0.32\textwidth]{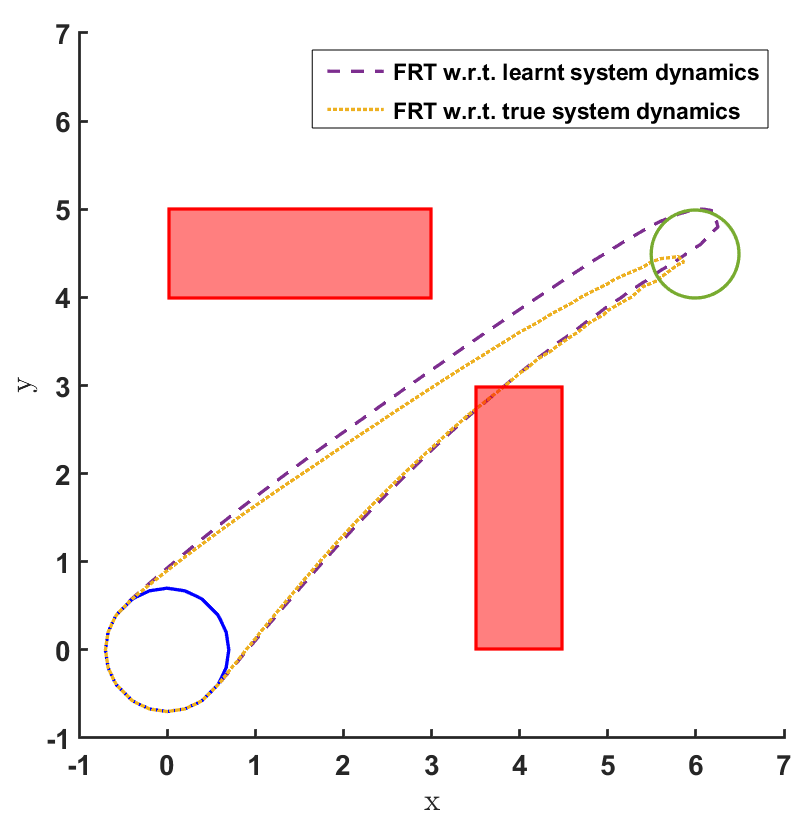}
\includegraphics[width=0.32\textwidth]{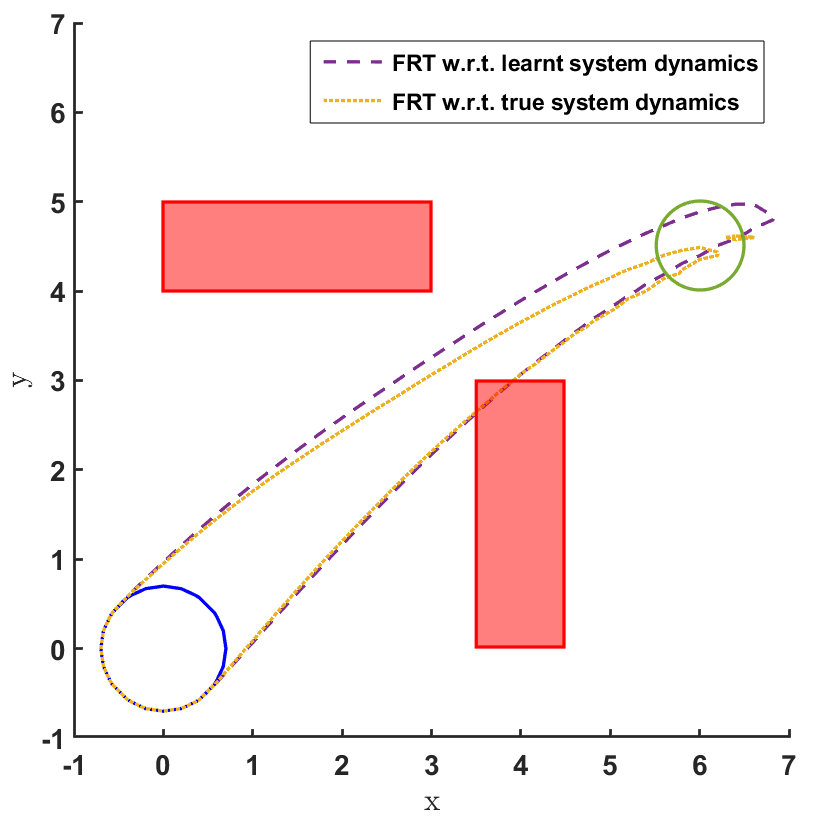}
\\
\includegraphics[width=0.32\textwidth]{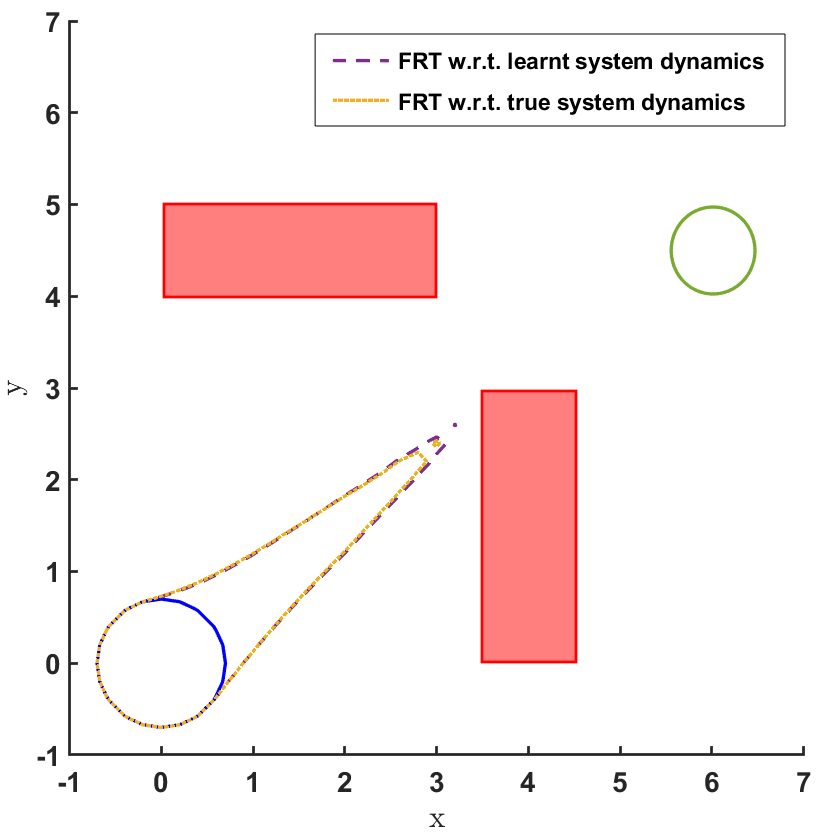}
\includegraphics[width=0.32\textwidth]{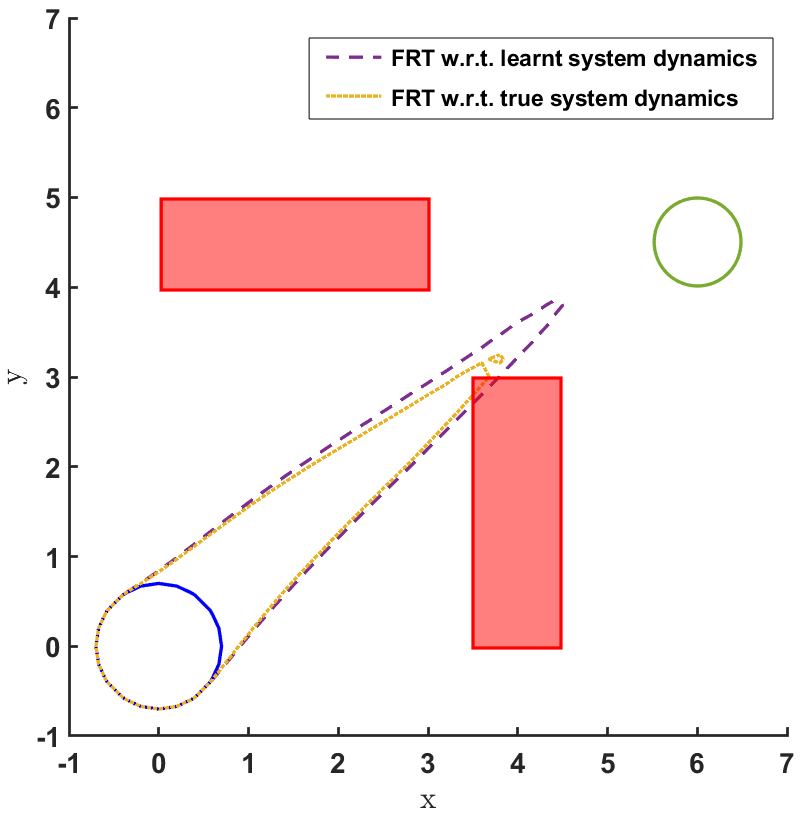}
\includegraphics[width=0.32\textwidth]{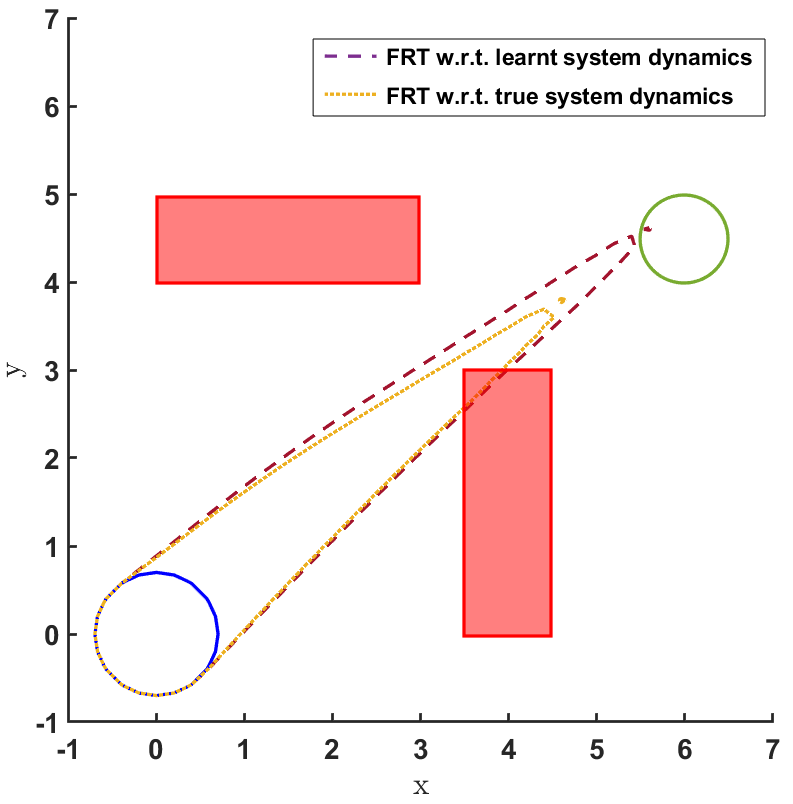}
\end{center}
\caption{The FRTs computed over the augmented learnt dynamics and the true system dynamics are presented. (Top Row) The FRTs corresponding to policy 1 (left), policy 2 (middle) and policy 3 (right) trained over learnt model 1. (Middle Row) The FRTs corresponding to policy 1 (left), policy 2 (middle) and policy 3 (right) trained over learnt model 2. (Bottom Row) The FRTs corresponding to policy 1 (left), policy 2 (middle) and policy 3 (right) trained over learnt model 3.}
\label{fig:FRT_2D}
\end{figure}

The FRT results for all the different models and controllers are shown in Fig. \ref{fig:FRT_2D}.
Each plot in Fig. \ref{fig:FRT_2D} shows two FRTs - the first corresponding to the augmented learnt dynamics and the second corresponding to the true system dynamics.
This is done to validate the result of the proposed safety verification framework with the ground truth classification of $\pi$ as safe or unsafe.
In general, it is observed that the FRT computed by the proposed framework, over the augmented learnt dynamics, closely resembles the FRT computed over the true system dynamics.
Thus, in every case, classification of $\pi$ determined by the proposed framework is consistent with the ground truth result.
All the above policies are found to be unsafe because the FRT intersects with obstacle 2.
In all the cases, the policies are safe with respect to obstacle 1.
Since all the above policies are unsafe, the BRT analysis is performed to determine $\sS_{safe}$ for each policy.

\begin{figure}[h]
\begin{center}
\includegraphics[width=0.32\textwidth]{images/BRT_300_100_v2.png}
\includegraphics[width=0.32\textwidth]{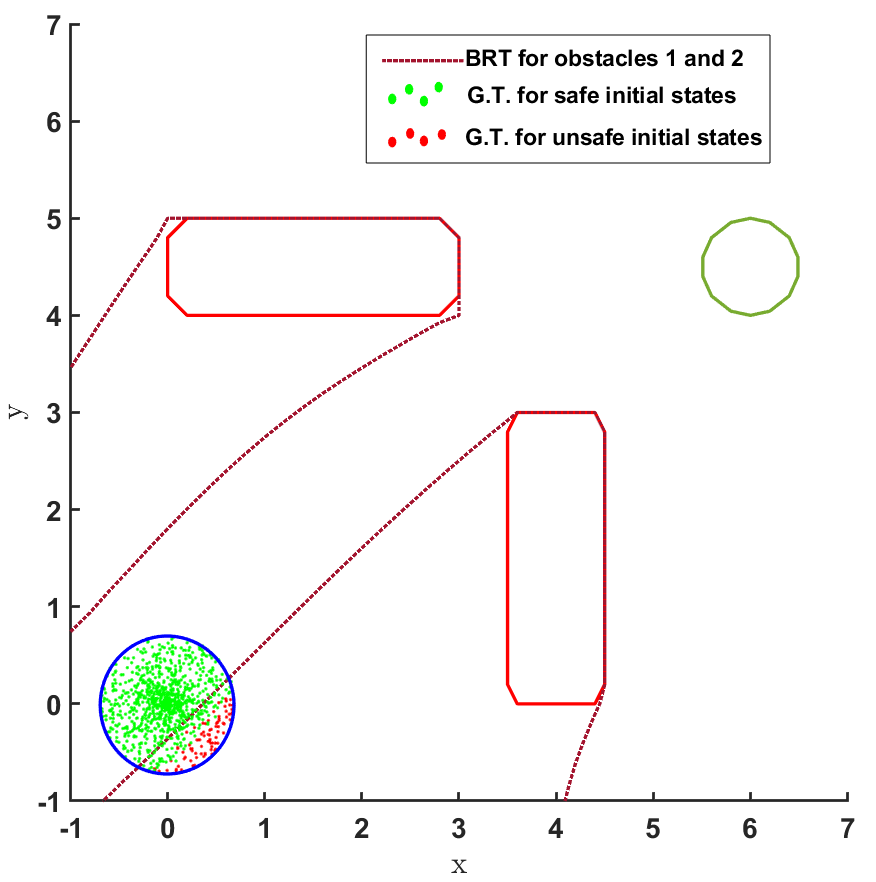}
\includegraphics[width=0.32\textwidth]{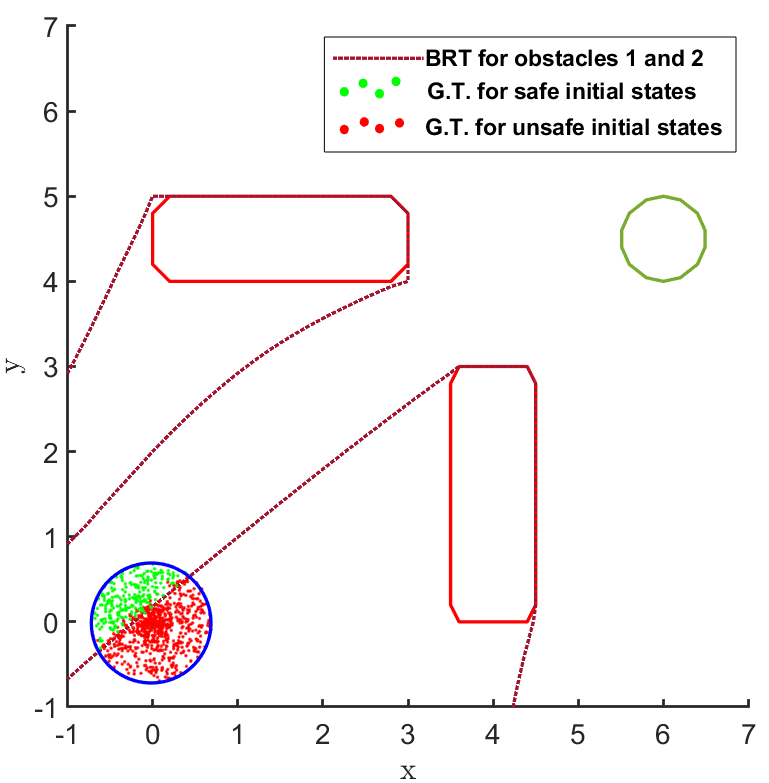}
\\
\includegraphics[width=0.32\textwidth]{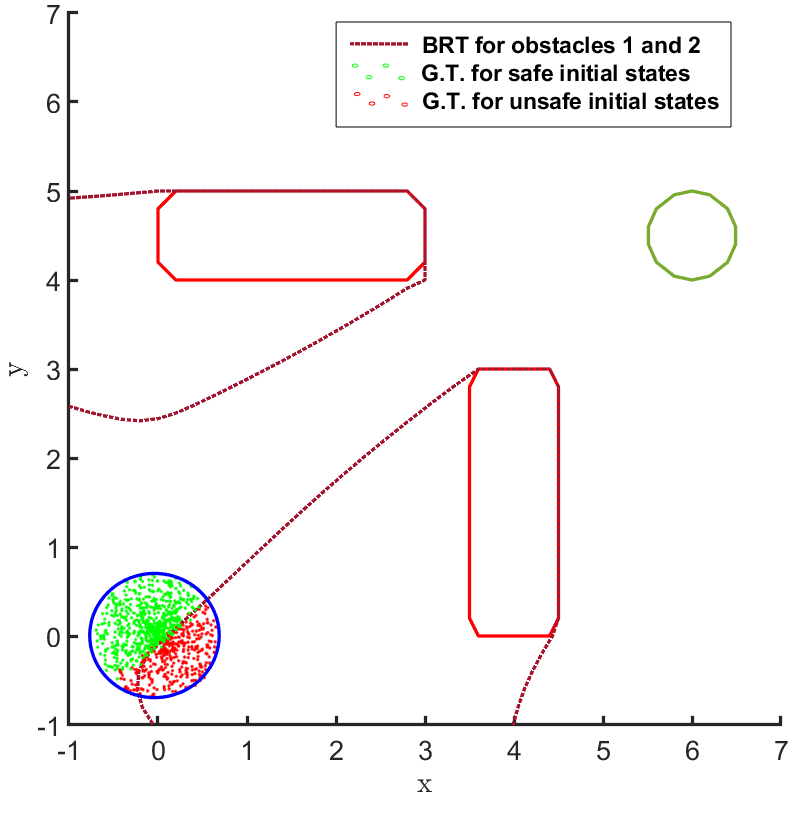}
\includegraphics[width=0.32\textwidth]{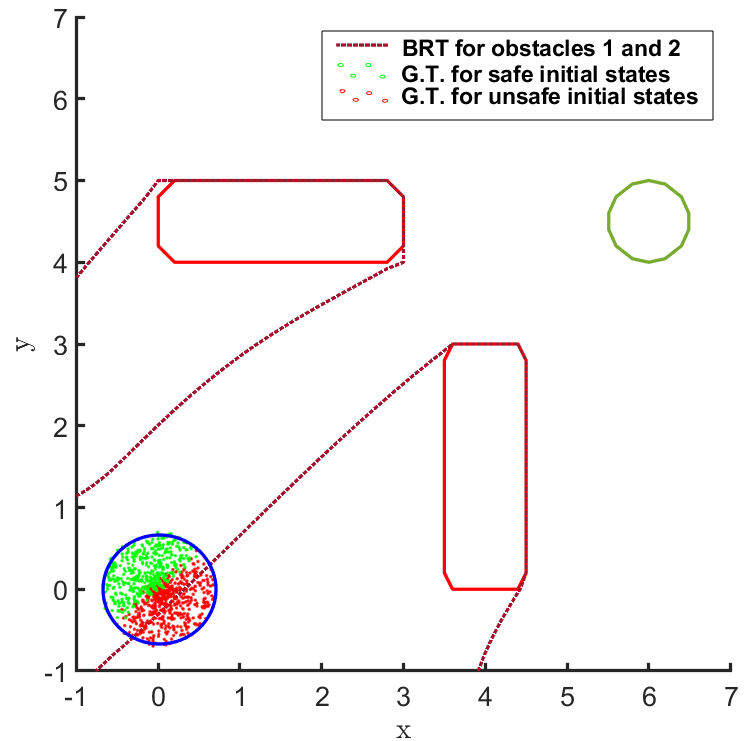}
\includegraphics[width=0.32\textwidth]{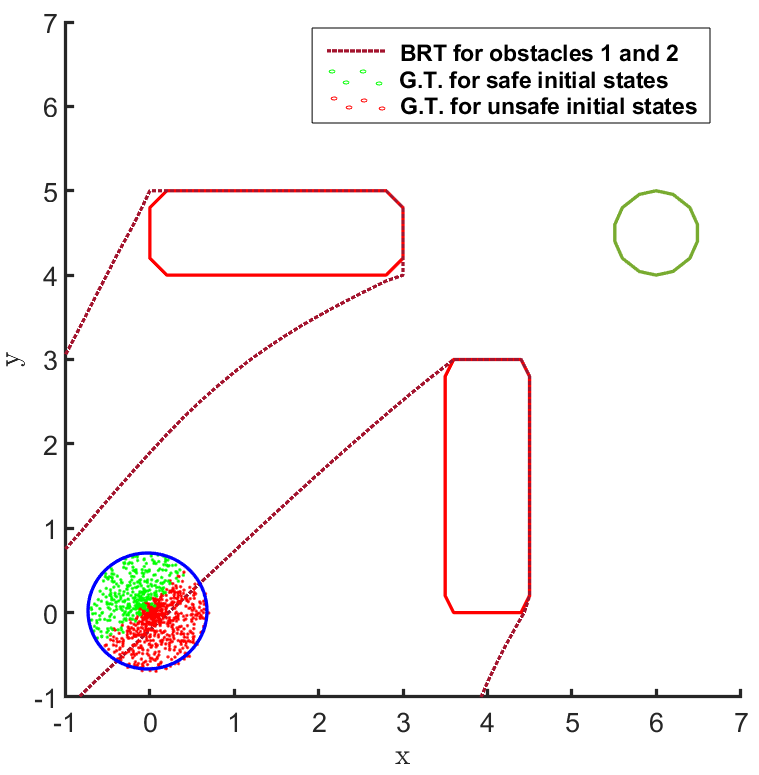}
\\
\includegraphics[width=0.32\textwidth]{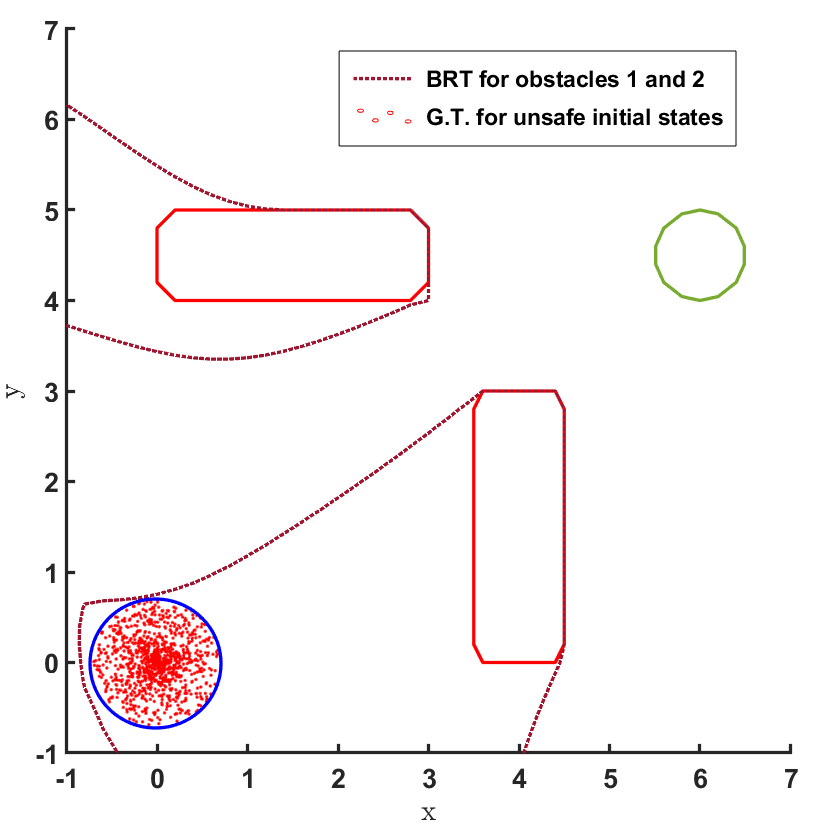}
\includegraphics[width=0.32\textwidth]{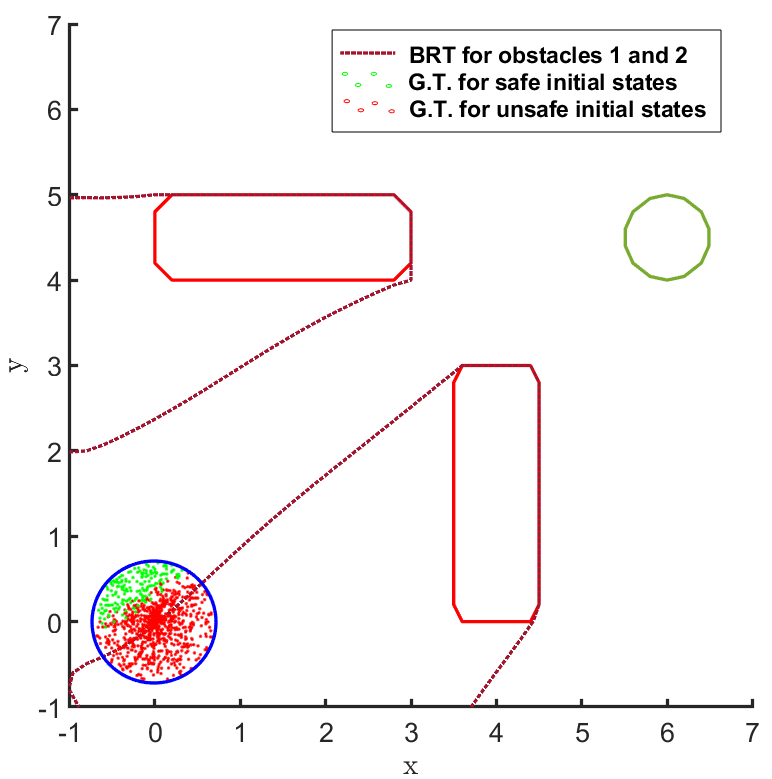}
\includegraphics[width=0.32\textwidth]{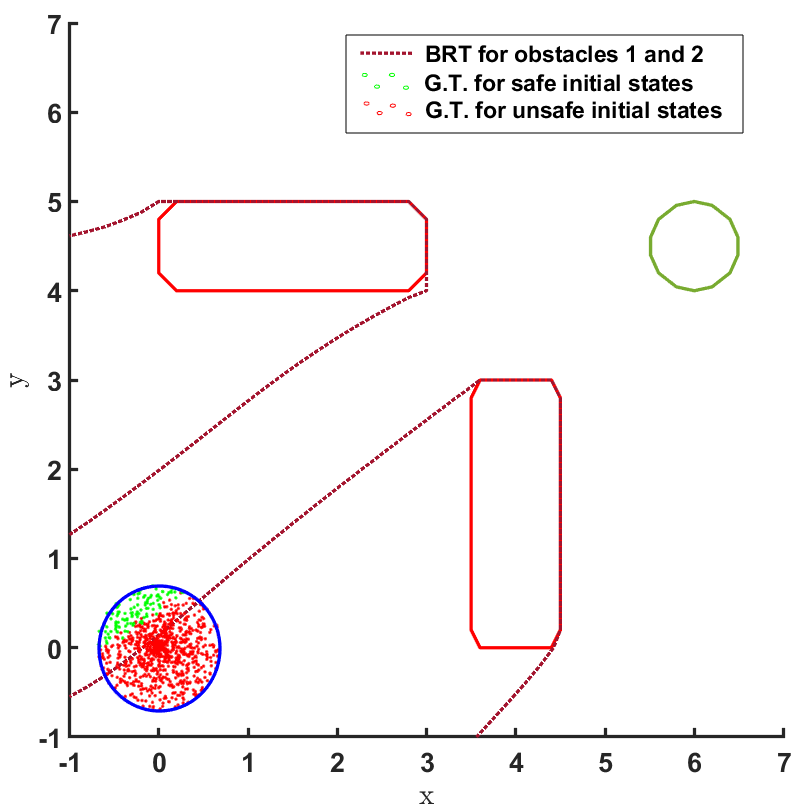}
\end{center}
\caption{ Comparison of $\sS_{safe}$ computed by the proposed BRT formulation with the ground truth (G.T.) data.  (Top Row) The BRT computed for policy 1 (left), policy 2 (middle) and policy 3 (right) trained over learnt model 1. (Middle Row) The BRT computed for policy 1 (left), policy 2 (middle) and policy 3 (right) trained over learnt model 2. (Bottom Row) The BRT computed for policy 1 (left), policy 2 (middle) and policy 3 (right) trained over learnt model 3.}
\label{fig:BRT_land_300}
\end{figure}

The results corresponding to the determination of $\sS_{safe}$, using the proposed BRT formulation, for each of the 3 controllers trained over learnt models 1, 2 and 3 are presented in Fig. \ref{fig:BRT_land_300}.
% , Fig. \ref{fig:BRT_land_600} and Fig. \ref{fig:BRT_land_1000}
%
$\sS_{safe}$ determined by the proposed method is compared with the ground truth (G.T.) data, which was generated by using Monte-Carlo samples of possible trajectories from $\sS_0$.
It is observed that the most accurate results are obtained for model 1.
For model 2, a small error due to modeling uncertainty is observed while determining $\sS_{safe}$.
This error due to modeling uncertainty is related to how conservative the bound on the modeling error is.
For model 3, the proposed BRT formulation correctly identifies policy 1 as completely unsafe, since $\sS_{safe} = \emptyset$.
For the remaining policies, a small error is observed while determining $\sS_{safe}$.

It can be seen from Fig. \ref{fig:BRT_land_300} that there is no consistent trend for the fraction of safe initial states across, either the models trained with increasing number of sample points (top to bottom) or, across the policies trained for increasing number of iterations (left to right).
This highlights one of the primary concerns for the lack of safety in RL based controllers: even though the cost function is designed to mitigate collisions and head towards the target, the controller learnt using it may not be safe because of various reasons stemming from state-representations learned by NN based policies and model dynamics, the hyper-parameters being used, or the optimization getting stuck in local minima, etc.
In such scenarios, the proposed framework can be used to check before deployment whether, the developed RL controller abides by the required safety constraints or not.

%
% \begin{figure}[h]
% \begin{center}
% \includegraphics[width=0.32\textwidth]{images/BRT_600_100_v2.png}
% \includegraphics[width=0.32\textwidth]{images/BRT_600_300_v2.png}
% \includegraphics[width=0.32\textwidth]{images/BRT_600_600_v2.png}
% \end{center}
% \caption{The $\sS_{safe}$ computed using the proposed BRT formulation for policy 1 (left), policy 2 (middle) and policy 3 (right) corresponding to model 2.}
% \label{fig:BRT_land_600}
% % \end{figure}
% %
% \begin{figure}[h]
% \begin{center}
% \includegraphics[width=0.32\textwidth]{images/BRT_1000_100_v2.png}
% \includegraphics[width=0.32\textwidth]{images/BRT_1000_300_v2.png}
% \includegraphics[width=0.32\textwidth]{images/BRT_1000_600_v2.png}
% \end{center}
% \caption{The $\sS_{safe}$ computed using the proposed BRT formulation for policy 1 (left), policy 2 (middle) and policy 3 (right) corresponding to model 3.}
% \label{fig:BRT_land_1000}
% \end{figure}

\subsection{Experiment Results: Safe Aerial Navigation}
\label{sec:results_aerial}
In the aerial navigation problem, an unmanned aerial vehicle (UAV) is flying in an urban environment.
To ensure a safe flight, the UAV has to avoid collision with obstacles, like buildings and trees, in it's path.
This problem is abstracted in a continuous state and action domain, where the state vector $\vs = [x, y, z]$ gives the 3D coordinates of the UAV and the action vector $\va = [v, \psi, \phi]$ comprises of the velocity, heading angle and pitch angle.
The UAV has to takeoff from the ground and reach a desired set of goal states.
Both the initial set of states and the goal states are represented using cuboids centered at coordinates $(0.0, 0.0, 0.0)$ and $(3.8, 4.5, 4.5)$, respectively.
There are two cylindrical obstacles in the environment, each of height $6$ units, which are centered at coordinates $(2.0, 4.0, 0.0)$ and $(4.0, 3.0, 0.0)$, respectively, with a radius of $0.5$ units.
This problem setting is presented in Fig. \ref{fig:3D_init}.

\begin{figure}[h]
\begin{center}
\includegraphics[width=0.32\textwidth]{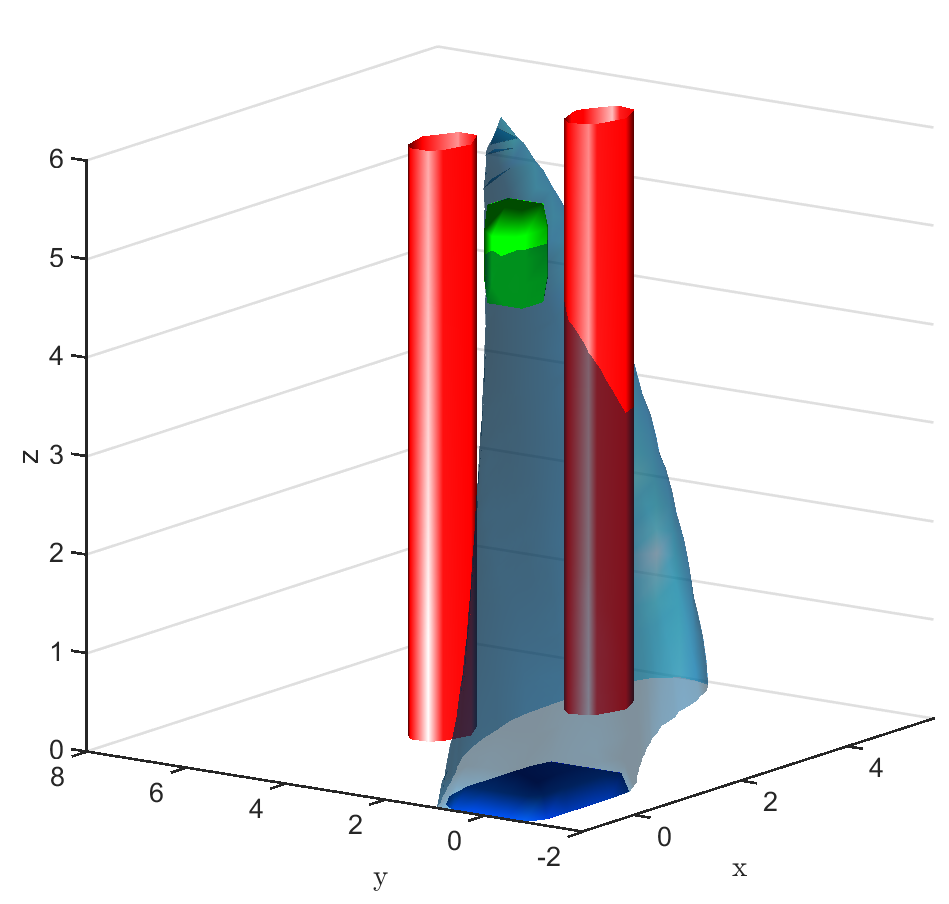}
\includegraphics[width=0.32\textwidth]{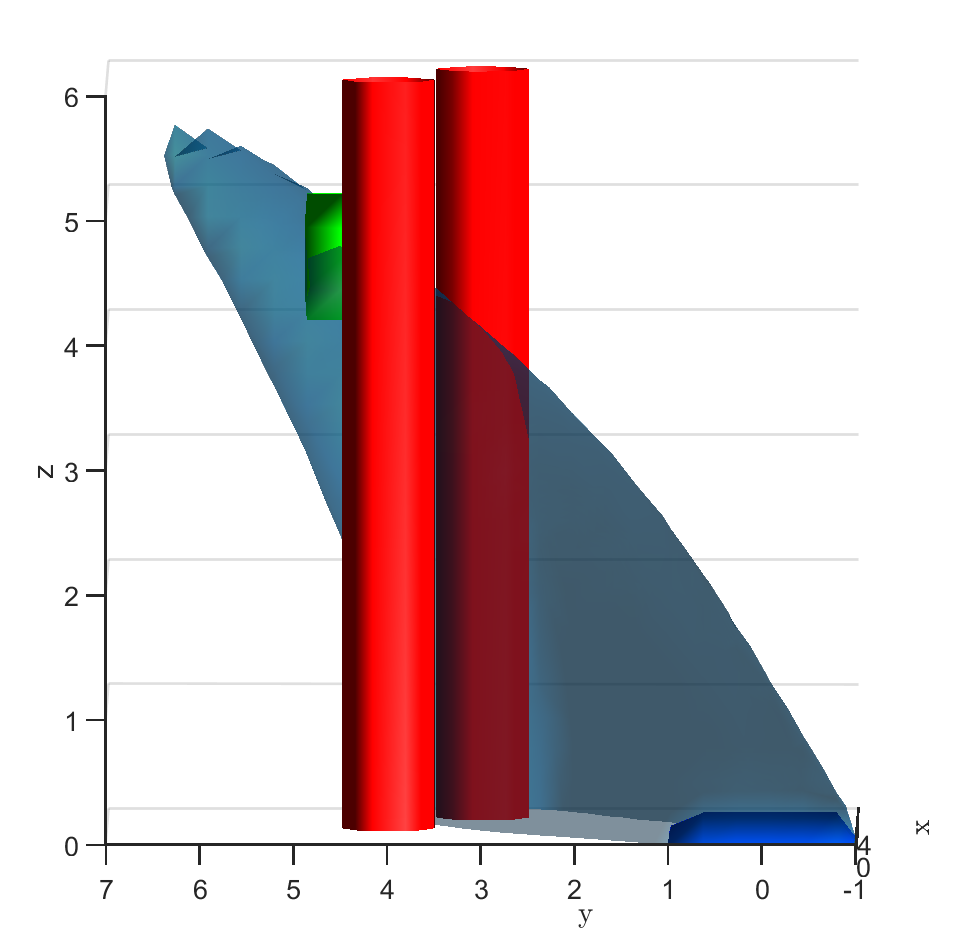}
\includegraphics[width=0.32\textwidth]{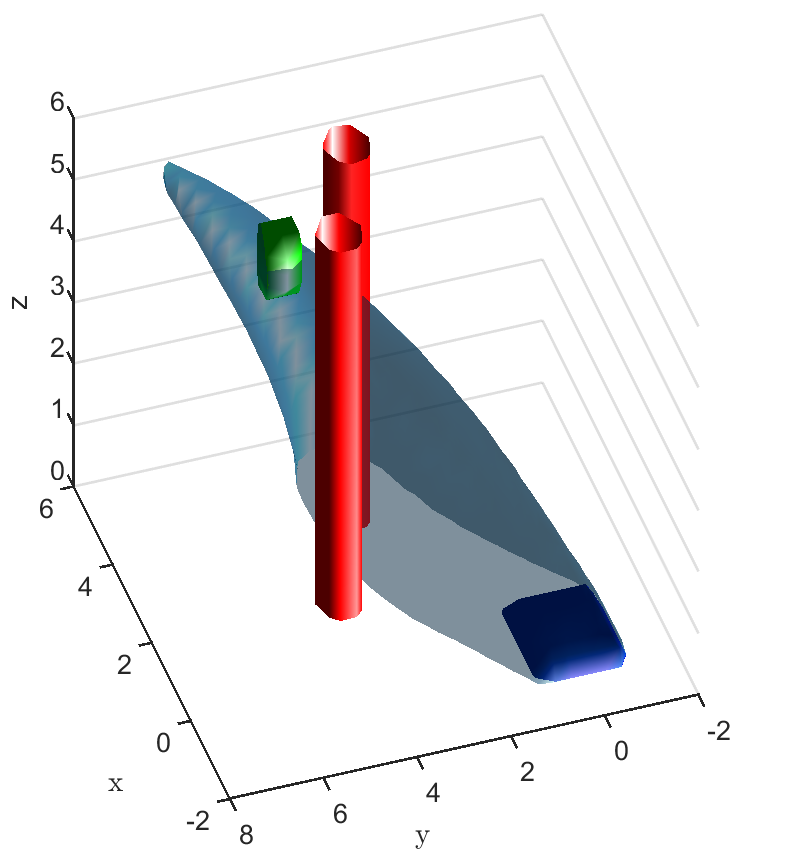}
\\
\includegraphics[width=0.32\textwidth]{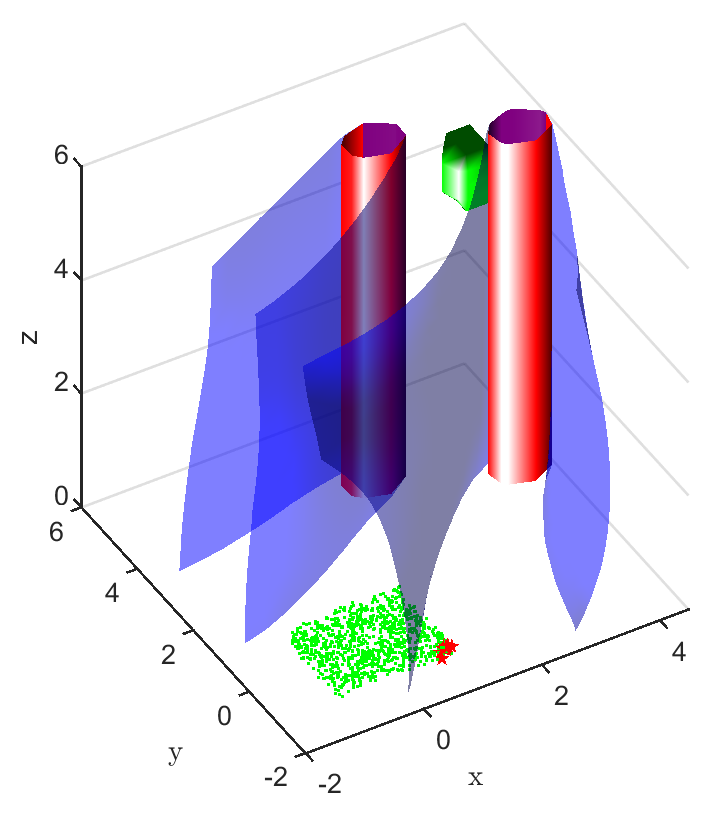}
\includegraphics[width=0.32\textwidth]{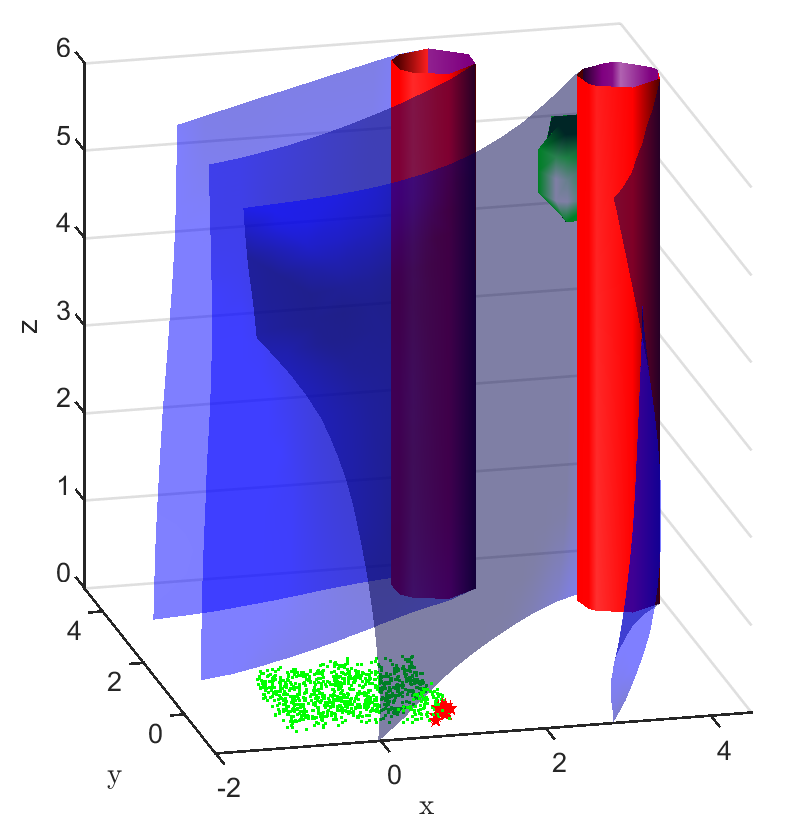}
\includegraphics[width=0.32\textwidth]{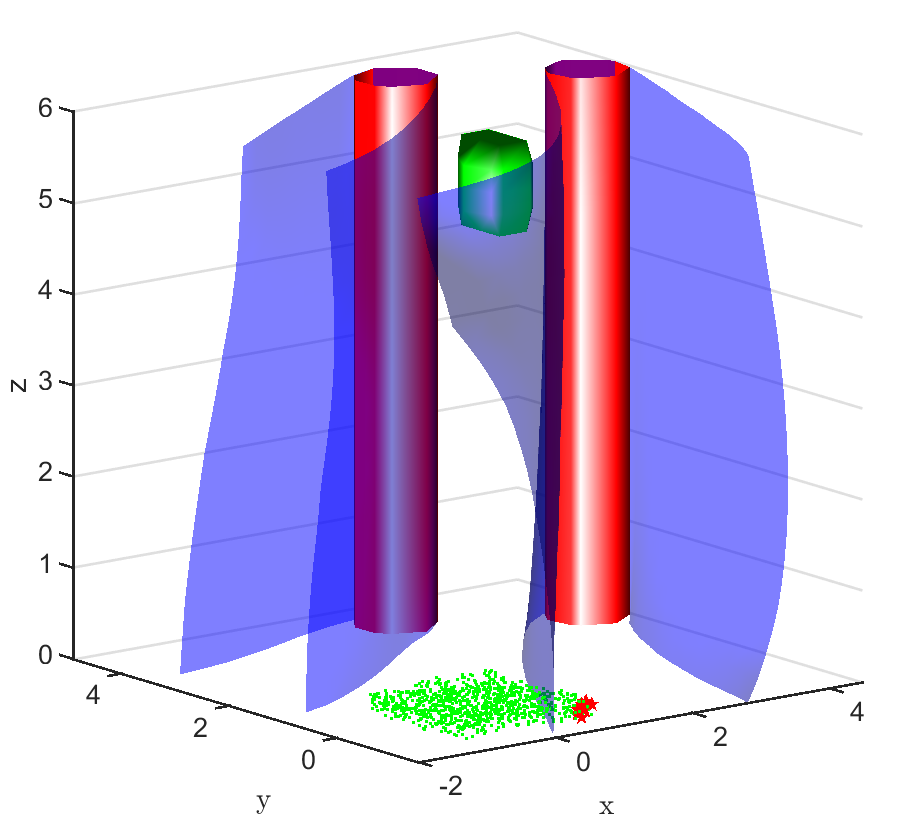}
\end{center}
\caption{(Top Row) Views from different angles for the FRT computed by the proposed method over the augmented learnt system dynamics.
It can be noticed that FRT intersects with obstacle 2, therefore, the policy is deemed unsafe.
(Bottom Row) Views from different angles of the BRT computed by the proposed method over the augmented learnt system dynamics.
The proposed method under approximates $\sS_{safe}$when compared with the G.T. data, marked in green. }
\label{fig:FRT_3D}
\end{figure}

A learnt model represented by an NN is trained over 1000 data points and the policy $\pi$ trained over this model is evaluated.
%
%The FRT computed by the proposed safety framework deems $\pi$ as unsafe because the FRT intersects with obstacle 2.
%
The FRT computed by the proposed framework, over the augmented learnt dynamics, is presented from different perspectives in Fig. \ref{fig:FRT_3D}.
While the computed FRT doesn't intersect with obstacle 1, it clearly intersects with obstacle 2.
Therefore, $\pi$ is deemed unsafe.
The next step is to compute the BRT and determine $\sS_{safe}$ for the unsafe policy $\pi$ as shown in Fig. \ref{fig:FRT_3D}.
It is observed that the proposed method under approximates $\sS_{safe}$ when compared with the G.T. data.

%% file: background_related_work.tex
\subsection{Extended Related Work} \label{sec:ERW}

% \todo{Intuition for reachable set. Convex vs non-convex discussion. Set of safe initial sets. }

% In this section, we briefly review the related work done so far in evaluating the properties of NNs.
%
NNs are one of the most commonly used function approximators for RL algorithms.
Therefore, often, verifying the safety of an RL controller reduces to verifying the safety of the underlying NN model or controller.
In this section, we briefly review the techniques used to evaluate the safety of NNs.

In recent years, NN-based supervised learning of image classifiers have found applications in autonomous driving and perception modules.
These are safety critical systems and it is necessary to verify the robustness of NN-based classifiers to minimize misclassification errors after they are deployed \citep{kwiatkowska2019safety}.
Some of the recent works have relied upon using existing optimization frameworks, such as \textit{linear programming} and \textit{satisfiability modulo theory} (SMT), to determine adversarial examples in the neighborhood of input images \citep{huang2017safety, bastani2016measuring, pulina2012challenging}.
However, these techniques usually 
% work for NNs with a small number of hidden nodes and often
require an approximation of the constraints on the NN model.
\citet{katz2017reluplex} improved the scalability of SMT verification algorithms and \citet{sun2019formal} used \textit{satisfiability modulo convex} (SMC) over a partitioned workspace, to determine the set of safe initial states for a robot, but their algorithms were limited to NNs with the $\texttt{ReLU}$ activation function.
However, the problem of interest in this work is more general than the \textit{single-step} classification problems considered in these related work, and instead we look at RL problems which require \textit{multi-step} decision making in real-valued domains. 

For real-valued output domains, the safety property needs to be verified for a set of states as opposed to a single label (i.e., a single point) in the output domain.
Therefore, to verify safety in the continuous problem domain, most works \citep{xiang2018reachable, tran2019parallelizable, xiang2018output, tran2019safety} first compute the forward reachable tube (FRT) of an NN, i.e, the set of output values achieved by the NN.
Then, the NN is said to be unsafe if the FRT intersects with an unsafe set.
\citet{xiang2018reachable}, \citet{tran2019parallelizable} and \citet{xiang2018output} follow the FRT algorithm by discretizing the input space into subspaces and executing forward pass over the NN to evaluate its output range.
However, the above works use convex polyhedrons to approximate non-convex reachable sets and the approximation error compounds over each layer of the NN.
\citet{tran2019safety} used a star-shaped polygon approach to generate the FRT which gives a less conservative solution than most polyhedra-based reachable tube computation.
Additionally, some of the algorithms are applicable specifically to $\texttt{ReLU}$ activation functions \citep{xiang2018reachable, tran2019parallelizable}.
In comparison, the algorithm proposed in this work computes the exact reachable tube and can handle most commonly used activation functions, including $\texttt{tanh, sigmoid, ReLu,}$ etc.